%% file: Main.tex
\documentclass[lettersize,journal]{IEEEtran}
\IEEEoverridecommandlockouts
\usepackage{cite}
\usepackage{amsmath,amssymb,amsfonts}
\usepackage[ruled]{algorithm2e}
\usepackage[dvipsnames,table]{xcolor}
\usepackage{graphicx}
\usepackage{pgfplots, tikz}
\usepackage{caption}
\usepackage{subcaption}
\usepackage{textcomp}
\usepackage{threeparttable}
\usepackage{multirow}
\usepackage{pifont}
\usepackage{listings}

\def\BibTeX{{\rm B\kern-.05em{\sc i\kern-.025em b}\kern-.08em
    T\kern-.1667em\lower.7ex\hbox{E}\kern-.125emX}}

\newcommand{\cmark}{\ding{51}}%
\newcommand{\xmark}{\ding{55}}%

\definecolor{revise}{RGB}{0, 0, 0}
\definecolor{revise2}{RGB}{0, 0, 0}

\pgfplotsset{compat=1.18}

\begin{document}

\RestyleAlgo{ruled}

\title{MiCoPro: End-to-End Mixed Precision HW/SW Co-design with HW-aware Proxy Model}

\author{
Zijun Jiang,~\IEEEmembership{Student Member,~IEEE,} and Yangdi Lyu,~\IEEEmembership{Member,~IEEE}
\thanks{Zijun Jiang, and Yangdi Lyu are with the Microelectronics Thrust, the Hong Kong University of Science and Technology (Guangzhou), Guangzhou 511453, China (email: zjiang438@connect.hkust-gz.edu.cn; yangdilyu@hkust-gz.edu.cn;).}
}

\markboth{Journal of \LaTeX\ Class Files,~Vol.~18, No.~9, September~2020}%
{How to Use the IEEEtran \LaTeX \ Templates}

\maketitle

\begin{abstract}
Quantized Neural Networks~(QNN) with low-bitwidth data have proven promising in efficient storage and computation on edge devices. To mitigate accuracy degradation while maximizing speedup, layer-wise mixed-precision quantization~(MPQ) becomes a popular solution. However, existing algorithms for exploring MPQ schemes are limited in flexibility and efficiency. Comprehending the complex impacts of different MPQ schemes on post-training quantization and quantization-aware training results is a challenge for conventional methods. Furthermore, an end-to-end framework for the optimization and deployment of MPQ models is missing in existing work.

To address these challenges, we propose the MiCo framework, a holistic MPQ exploration and deployment framework for edge AI applications. The framework adopts a novel optimization algorithm to search for accuracy-optimal quantization configurations under strict latency constraints. We further extended the framework to MiCoPro, which introduces a robust Hardware-Aware Proxy (HAP) model to enhance prediction accuracy and hardware versatility. By leveraging target-specific latency modeling, MiCoPro enables rapid exploration and direct deployment from PyTorch models to bare-metal C code. 
We demonstrate the versatility of our framework on both the BitFusion accelerator and SIMD-extended RISC-V processors, achieving up to 40\% of latency reduction with less than 3\% of accuracy drop. 

\end{abstract}


\input{Content/Introduction}

\input{Content/Background}
\input{Content/RelatedWork}

\input{Content/MPQ_Search}

\input{Content/Hardware_Modeling}

\input{Content/EndToEnd_Deployment}

\section{Experimental Results}

\subsection{MPQ Search Setup}

To demonstrate the generalizability of our approach, we extend our evaluation beyond the CNN-based image classification models commonly used in prior works~\cite{edge-mpq,haq,hawqV3}. Instead, we evaluate across a highly diverse set of tasks, architectures, and scales, including:
\begin{itemize}
    \item \textbf{Vision}: CNN4 \& VGG7 (CIFAR-10~\cite{cifar}), LeNet5 (MNIST), ResNet-18 \& SqueezeNet (CIFAR-100~\cite{cifar}), ResNet-34 \& ViT (ImageNet~\cite{imagenet}).
    \item \textbf{Audio}: DS-CNN~(Speech Commands V2~\cite{speechcmd}).
    \item \textbf{Language}: TinyLLaMa-1M \& -7M~(TinyStories~\cite{tinystories}).
\end{itemize}

Tab.~\ref{tab:networks} lists the various baseline models used in our experiments, sorted by their number of layers.

\begin{table}[b]
\centering
\begin{threeparttable}
    \caption{Baseline Full Precision Models}
    \begin{tabular}{| c | c | c | c |}
    \hline
    \textbf{Model} & \textbf{Acc.~(\%)} & \textbf{\# Layers}     & \textbf{MACs}\\
    \hline
    CNN4~\cite{cmsis-cnn}          & 75.79  & 4 & 12.3M        \\
    \hline
    LeNet5~\cite{lenet}            & 99.16  & 5 & 0.39M         \\ 
    \hline
    VGG7~\cite{vgg}     & 83.13 & 7 & 110.9M \\
    \hline
    DS-CNN~\cite{dscnn}  & 88.72 & 10 & 29.0M \\
    \hline
    ResNet-18~\cite{resnet} & 71.40 & 21 & 565.1M \\
    \hline
    SqueezeNet~\cite{squeezenet}   & 68.60  & 26  & 53.3M        \\
    \hline
    TinyLLaMa-1M~\cite{llama2c}    & 60.41\tnote{*}  & 36  & 1.35M \\
    \hline
    ResNet-34~\cite{resnet} & 73.30 & 37 & 3.68G \\
    \hline
    ViT-B-32~\cite{vision_transformer} & 75.91 & 38 & 180.1M \\
    \hline
    TinyLLaMa-7M~\cite{llama2c}    & 67.71\tnote{*}  & 43  & 7.15M \\
    \hline
    \end{tabular}
    \label{tab:networks}
\begin{tablenotes}
\item [*] For TinyLLaMa, the accuracy refers to next token prediction accuracy. 
\end{tablenotes}
\end{threeparttable}
\end{table}


The training and testing of models are conducted on a server with an Nvidia A30 GPU, and all of the models and quantization are implemented with PyTorch.

We compared our method with three state-of-the-art methods mentioned in Tab.~\ref{tab:SW_Cmp}, which include the NLP-based $w$-method (Edge-MPQ)~\cite{edge-mpq}, the RL-based HAQ method~\cite{haq}, and the BO-based BOMP~\cite{van2023bomp}. To facilitate this comparison, we directly integrated the open-source implementation of HAQ into our framework and re-implemented the other two methods based on their published descriptions. \textcolor{revise}{Additionally, we evaluate the ILP-based HAWQ-V3~\cite{hawqV3} on representative models, as detailed in Sec.~\ref{sec:uniform_baseline}. We do not include gradient-based differentiable methods (e.g., \cite{SDQ, bayesian_bits}) in the comparison, as their joint optimization of network parameters and quantization choices requires a dedicated end-to-end training procedure, which is prohibitively time-consuming compared to other methods.}

For our method and two machine learning-based methods~(HAQ and BOMP), the sample budgets are set to the same numbers for fair comparisons. In the case of the $w$-based method, which performs static analysis of layer-wise sensitivities, the required number of samples equals the number of layers in each model.

We conduct the exploration under different BOPs constraints, which are defined as specific ratios of the base BOPs derived from INT8 models, where all $b_w$ and $b_a$ values are fixed at 8.

\input{Content/PTQ_Experiment}

\input{Content/QAT_Experiment}

\input{Content/NewExperiment}

\subsection{Ablation Study}

To validate the effectiveness of our proposed techniques, we conduct ablation experiments on the PTQ search problem for SqueezeNet and TinyLLaMa-1M under a 0.6 BOPs constraint with 32 search budgets (16 initial + 16 search). Starting with the basic random forest models, we incorporate the two main sampling techniques discussed in Sec.~\ref{sec:initial} and Sec.~\ref{sec:optimization} one by one to assess their impact.

\begin{table}[b]
    \centering
    \caption{Ablation Study on PTQ Search}
    \begin{tabular}{l|l|l}
    \hline
    \textbf{Method}      & \textbf{SqueezeNet Acc.} (\%) & \textbf{TinyLLaMa Acc.} (\%)\\
    \hline
    Random Forest        &  66.95                        & 55.92 \\
    + Orthogonal Initial &  67.13 (+0.18)                & 56.66 (+0.74) \\
    + Near-Constraint    &  \textbf{67.92} (+0.79)        & \textbf{57.89} (+1.23)\\
    \hline
    \end{tabular}
    \label{tab:ablation}
\end{table}

As shown in Tab.~\ref{tab:ablation}, the optimization results are first improved by the orthogonal initial samples, which enhances the initial learning of the accuracy predictors. Then, by focusing on the near-constraint spaces, the optimization results are further promoted.

\subsection{Overall Runtime Analysis}

To characterize the computational efficiency of our framework, we investigate the runtime distribution across the exploration flow. Our analysis indicates that the computational bottleneck resides primarily in the model evaluation stage rather than the search algorithm itself. While the searching algorithm encompasses several steps—such as near-constraint sampling and predictor model training—these operations account for a disproportionately small percentage of the overall runtime.

We exemplify this with a SqueezeNet case study: while PTQ evaluation requires 1.46s per iteration, our algorithm incurs only 0.33s of overhead—comparable to baseline methods like HAQ (0.16s) and BO (0.12s). The disparity is further highlighted in QAT scenarios, where evaluation time extends to 5 minutes per sample, making the search overhead mathematically negligible.

\section{End-to-End MiCo Exploration Results}

As introduced in Sec.~\ref{sec:deployment_flow}, the MiCo framework is capable of deploying the MPQ models explored to various hardware targets in a hardware-aware fashion. To demonstrate this hardware-awareness, we run the flows with both the conventional hardware-agnostic BOPs constraint and our hardware-aware proxy model constraint to show the impact of proxy models on the final hardware latencies.

\subsection{Mixed Precision Hardware Targets}

To demonstrate the end-to-end exploration and deployment flow of our MiCo framework, we select two hardware platforms as showcases and run the complete flows on them for MPQ models. The two platforms are the aforementioned mixed-precision accelerator and the extended CPU in Fig.~\ref{fig:corr_bops}, and their detailed descriptions are as follows.

\subsubsection{BitFusion}
The BitFusion~\cite{bit_fusion} accelerator is a bit-flexible systolic array accelerator, which supports mixed-precision computations from 2-bit to 8-bit integers. For the first showcase of a complete MiCo flow, we use the BitFusion simulator to set up an accelerator with 32$\times$16 processing elements as the hardware target for our end-to-end exploration. A batch size of 16 is assumed for the inference on BitFusion.

\color{revise}
\subsubsection{VexiiMiCo}

VexiiMiCo extends the 32-bit VexiiRiscv CPU~\cite{vexiiriscv}
with ten single-cycle R-type SIMD dot-product instructions, whose
detailed design was introduced in our conference
work~\cite{mico_iccad}. The extension supports ten combinations of
8/4/2/1-bit operands packed into two 32-bit source registers, allowing
each register to contain $4\times$INT8, $8\times$INT4,
$16\times$INT2, or $32\times$INT1 elements. It also supports
asymmetric-precision dot products by extending the lower-precision
operands before multiplication.

On a $64\times64\times64$ MatMul benchmark, VexiiMiCo achieves a
$3.7\times$ speedup for INT8 and up to $19.2\times$ for lower
precisions over the baseline RV32IMFC CPU~\cite{mico_iccad}. To
evaluate MiCoPro across different processor configurations, we use
three variants sharing the same mixed-precision extension:
\textit{Tiny}, a single-issue core without caches or branch prediction;
\textit{Small}, a single-issue core with I/D caches and a Branch Target
Buffer~(BTB); and \textit{High}, a dual-issue core with larger caches
and a gshare branch predictor.
\color{black}

\subsection{End-to-End Proxied Latency Prediction Accuracy}

\subsubsection{Hardware Profile Creation}
\label{sec:dataset_size}

As noted in Sec.~\ref{sec:profiling}, constructing target-specific latency proxy models requires kernel-profiling datasets. We collected the kernel profiling data with different precisions and shapes on both hardware platforms. Table~\ref{tab:dataset size} summarizes the total number of kernel latency samples collected for each target.

\begin{table}[h]
    \centering
    \caption{The Number of Kernel Latency Samples}
    \begin{tabular}{l|l|l}
    \hline
    \textbf{Hardware} & \textbf{MatMul} & \textbf{Conv2D}\\
    \hline
        BitFusion                   & 342 & 2400 \\
        VexiiMiCo-\textit{Tiny}     & 98  &  735 \\
        VexiiMiCo-\textit{Small}    & 98  &  735 \\
        VexiiMiCo-\textit{High}     & 98  &  735 \\
    \hline
    \end{tabular}
    \label{tab:dataset size}
\end{table}
The sampled data is richer on BitFusion thanks to its fast Python-based simulation. \textcolor{revise}{On VexiiMiCo CPUs, the slower RTL simulation yields less data and is more time-consuming to collect.} As previously mentioned, the process is a one-time effort: once enough data is collected, it can be used in all future MPQ exploration.


\subsubsection{Latency Prediction with Min-Max Calibration}
\label{sec:lat_prediction}
To evaluate the accuracy of the proposed proxy modelling method for latency prediction, we randomly sample different MPQ schemes on different networks and evaluate them on different hardware targets to get the latency results. For VGG7 and ResNet18 on the BitFusion target, 64 schemes are evaluated per network. For CNN4 on the VexiiMiCo Small and High targets, 32 schemes are evaluated per target. XGBoosting Regressor is used as the predictor for the BitFusion target, and Random Forest Regressor is used as the predictor for the VexiiMiCo targets.

We mainly evaluate two metrics for the proxy models: mean absolute percentage error~(MAPE) and coefficient of determination~($R^2$). The evaluation results are listed in Tab.~\ref{tab:proxy_bf} and Tab.~\ref{tab:proxy_mico}. \textcolor{revise}{The feature sets compared in the tables are BOPs, CBOPs, and ``Full'' features (denoted as Full Feat.), as detailed in Sec.~\ref{sec:features}.}

\input{ProxyTable}

The results demonstrate the effectiveness of our proxy models across
different networks and hardware targets. Using BOPs alone reduces
latency prediction to a one-dimensional mapping and overlooks
architecture-dependent effects. Although min--max calibration maintains
acceptable MAPE, the resulting correlation remains poor; in particular,
BOPs produces negative $R^2$ values for all VexiiMiCo variants in
Tab.~\ref{tab:proxy_mico}. CBOPs substantially improves both metrics by
decomposing hardware cost into BMACs, ALoads, and WLoads, thereby
capturing operand-dependent memory traffic and compute parallelism.
Further incorporating the ``Full'' kernel features introduced in
Sec.~\ref{sec:features} enables non-linear regressors such as XGB and RF
to achieve the most reliable latency predictions.




\begin{figure}[t]
    \centering
    \begin{subfigure}[b]{0.24\textwidth}
        \centering
        \includegraphics[width=\textwidth]{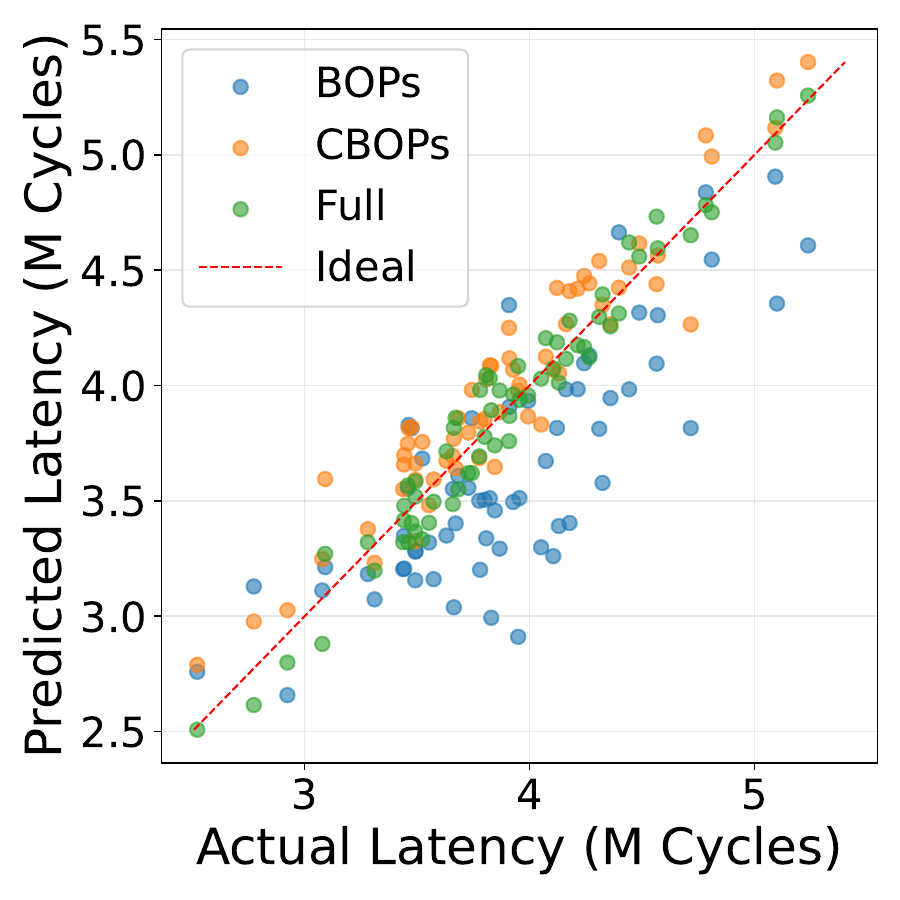}
        \caption{VGG7 on BitFusion}
        \label{fig:subfig_a}
    \end{subfigure}
    \hfill
    \begin{subfigure}[b]{0.24\textwidth}
        \centering
        \includegraphics[width=\textwidth]{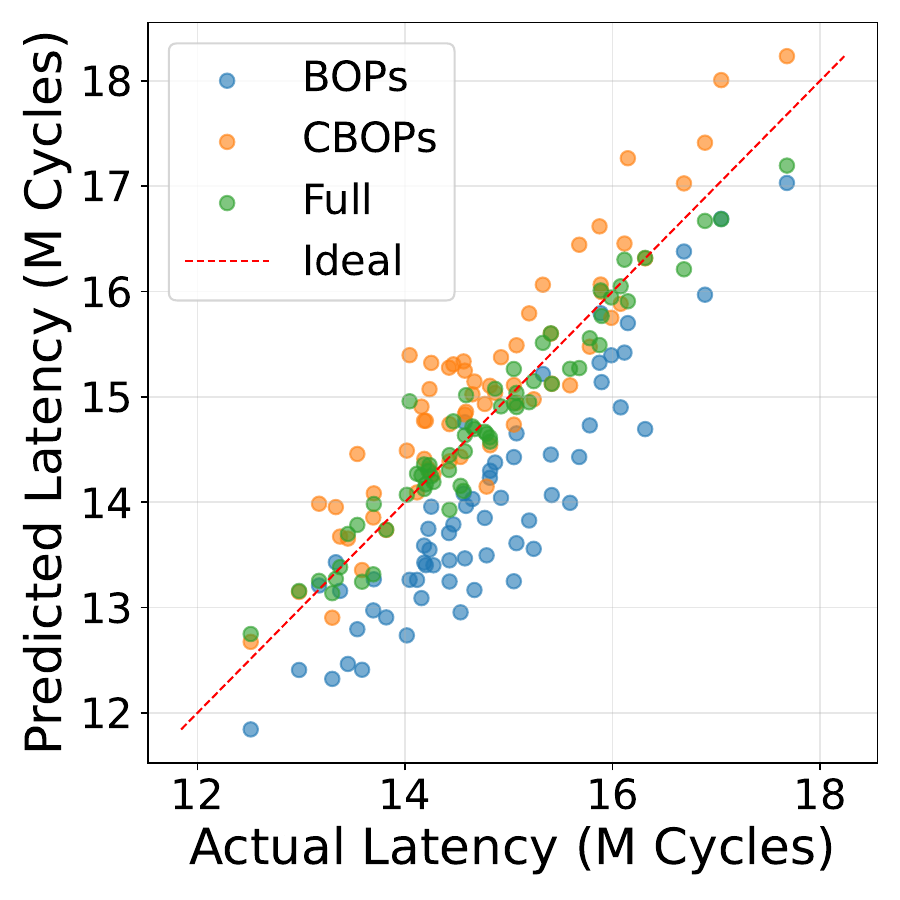}
        \caption{ResNet-18 on BitFusion}
        \label{fig:subfig_b}
    \end{subfigure}
    \begin{subfigure}[b]{0.24\textwidth}
        \centering
        \includegraphics[width=\textwidth]{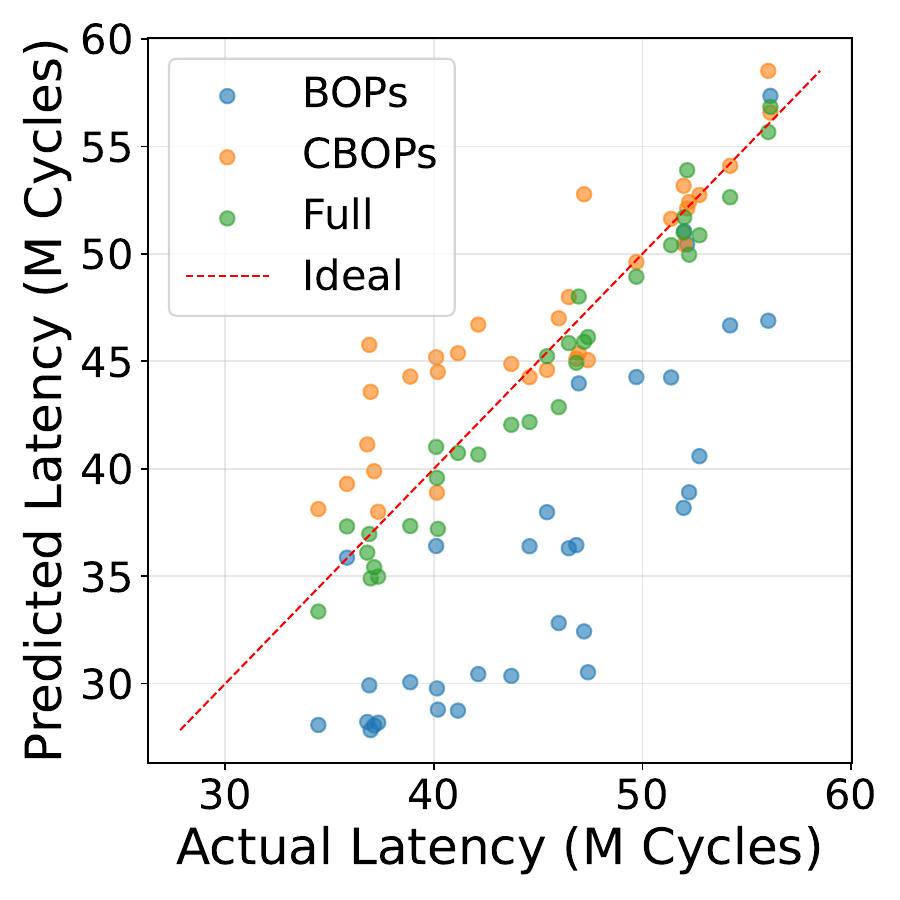}
        \caption{CNN4 on VexiiMiCo-Small}
        \label{fig:subfig_c}
    \end{subfigure}
    \hfill
    \begin{subfigure}[b]{0.24\textwidth}
        \centering
        \includegraphics[width=\textwidth]{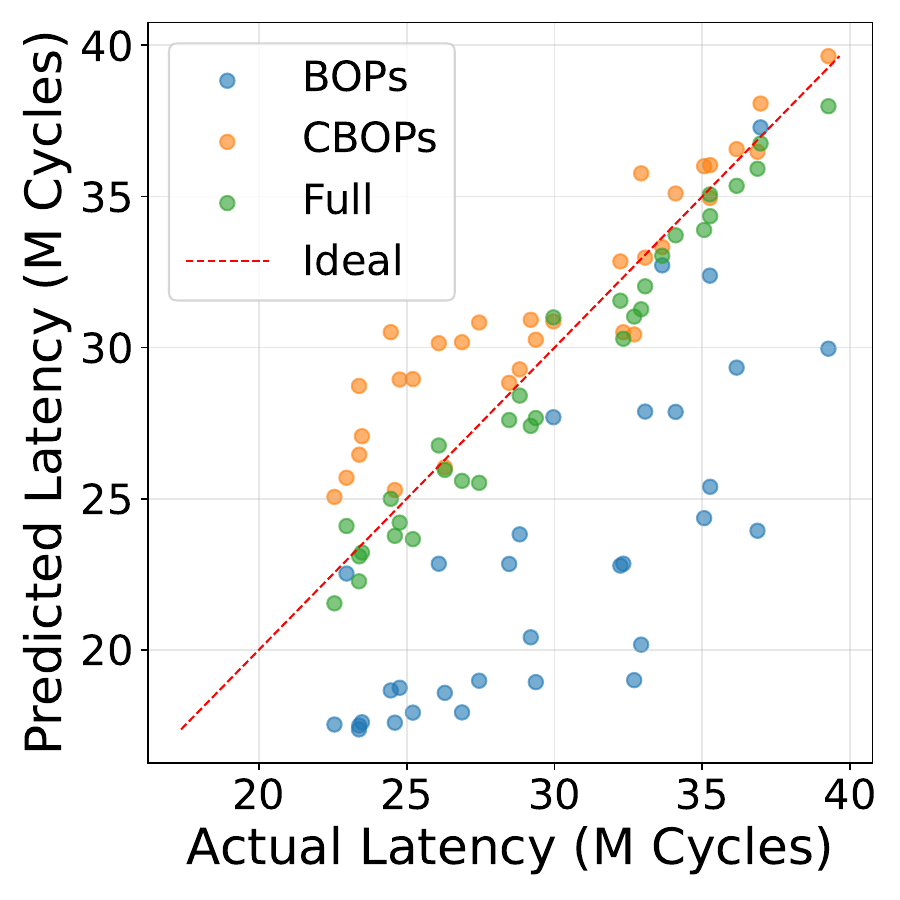}
        \caption{CNN4 on VexiiMiCo-High}
        \label{fig:subfig_d}
    \end{subfigure}

    \caption{End-to-End Latency Prediction with Proxy Models}
    \label{fig:proxy_acc}
\end{figure}

We also visualize the prediction scatter in Fig.~\ref{fig:proxy_acc} for a more intuitive comparison. It can be clearly observed that the CBOPs-based and \textcolor{revise}{full kernel feature-based models (denoted as Full)} produce better predictions. BOPs-based models, however, generally lead to underestimations, which can be seen from the fact that most blue points are under the ideal prediction line in the above figures, which matches the undesirable MAPE and $R^2$ results in the tables.

\color{revise}
\subsubsection{CBOPs Component Ablation}
To further analyze the contribution of each CBOPs component, we ablate the CBOPs composition by comparing BOP, the($(b_w+b_a)\cdot\text{MACs}$, BMACs alone, and the full CBOPs $\{$BMACs, ALoads, WLoads$\}$, each fit with linear regression and averaged over the same five prediction tasks. The results are listed in Tab.~\ref{tab:cbops_ablation}.
\begin{table}[b]
\color{revise}
\centering
\caption{Ablation of CBOPs Components}
\label{tab:cbops_ablation}
\begin{tabular}{l | c c}
\hline
Feature & MAPE$\downarrow$ & \textbf{$R^2$}$\uparrow$ \\
\hline
BOPs & 18.02 & 0.3790 \\
$(b_w+b_a)\cdot\text{MACs}$ & 9.30 & 0.7451 \\
BMACs & 20.90 & 0.3988 \\
\textbf{CBOPs} & \textbf{6.50} & \textbf{0.8432} \\
\hline
\end{tabular}
\end{table}
These results confirm the rationality and effectiveness of CBOPs: decomposing the hardware cost into the max-based compute term and the separate activation/weight load terms yields a markedly more accurate latency proxy.
\color{black}

\color{revise}
\subsubsection{Proxy Accuracy-Speed Trade-off}

For a more comprehensive study on the trade-off of feature selection and regressor model selection, 
we use the same 5 sets of prediction tasks of Sec.~\ref{sec:lat_prediction} above to evaluate the prediction quality and time per scheme of different feature/regressor combinations.
\begin{table}[t]
\color{revise}
\centering
\caption{Proxy Accuracy-Speed Trade-off}
\label{tab:proxy_tradeoff}
\begin{tabular}{l | l | c c c}
\hline
 Feature & Regressor & MAPE$\downarrow$ & $R^2\uparrow$ & Time (ms) \\
\hline
CBOPs & Linear & 6.50 & 0.84 & \textbf{0.11} \\
CBOPs & XGB & 5.90 & 0.78 & 0.45 \\
CBOPs & RF  & 6.43 & 0.68 & 5.91 \\
\hline
Full Feat. & Linear & 8.34 & 0.52 & \textbf{0.11} \\
Full Feat. & XGB & 3.96 & 0.84 & 0.44 \\
Full Feat. & RF  & \textbf{3.87} & \textbf{0.85} & 5.63 \\
\hline
\end{tabular}
\end{table}
Tab.~\ref{tab:proxy_tradeoff} reports the inference time per prediction (ms). CBOPs + linear regression achieves strong accuracy at the lowest cost, while full features require non‑linear models for top accuracy but at higher cost, which aligns with the trade‑off discussed in Sec.~\ref{sec:features}. 
This proxy inference overhead mainly affects candidate screening during near-constraint sampling: it can be noticeable for small networks with inexpensive evaluations, but becomes marginal for larger networks.
\color{black}

\subsubsection{Latency Prediction with Inter-Hardware Transfer}
As mentioned in Sec.~\ref{sec:dataset_size}, many kernel profiles need to be collected, and the process could be time-consuming with slow simulation.

To evaluate the efficiency of our transfer learning method, we first fit the full feature-based proxy models with random forest regressors directly on a small proportion~(5\%) of profile data as the baseline. For comparison, we apply transfer learning by transferring from fully trained proxy models on other VexiiMiCo variants to the target variant, which has the same small dataset. The results are listed in Tab.~\ref{tab:transfer_learning}.

\input{TransferTable}

The transfer learning largely improves the quality of the latency prediction on new target hardware with minimal new profile data, especially for the $R^2$, which is crucial for the constraint-based MPQ exploration. \textcolor{revise}{For our data collection process, gathering profiling data for a single VexiiMiCo variant takes 27 hours to complete, while transferring to another variant with 5\% of new data costs less than 2 hours, which greatly saves the profiling time required for the new variants.}

\subsection{End-to-End Exploration on BitFusion Accelerator}

\begin{figure*}[htbp]
\begin{subfigure}[b]{0.41\textwidth}
\input{vgg_bitwidths}
\end{subfigure}
\begin{subfigure}[b]{0.58\textwidth}
\input{resnet18_bitwidths}
\end{subfigure}
\caption{Explored MPQ Scheme on BitFusion under 60\% Latency Constraint}
\label{fig:bf_schemes}
\end{figure*}


With the MiCo framework, we conduct MPQ exploration on VGG and ResNet architectures and deploy the resulting models on the cycle-accurate BitFusion simulator to evaluate actual hardware latencies. The exploration budget is set to 20 samples (10 initial samples + 10 search samples).


\begin{table}[ht]
\centering
\caption{Exploration \& Deployment on BitFusion}

\label{tab:vgg_on_bf}
\begin{tabular}{l | lll}
\hline
\multicolumn{4}{c}{\textbf{VGG7 (CIFAR-10)}} \\
\hline
\textbf{Precision} &\textbf{Acc.} (\%) & \textbf{Constraint} & \textbf{Cycles} (Ratio) \\
\hline
8-bit     & 82.97 & -        & 6.26M (1.0$\times$)  \\
Mixed     & 83.14 & 0.6$\times$BOPs  & 4.48M (\textbf{\textcolor{BrickRed}{0.72$\times$}}) \\
Mixed     & 82.49 & 0.6$\times$\textit{Proxy} & \textbf{3.73M} (\textbf{\textcolor{ForestGreen}{0.60$\times$}}) \\
\hline
\multicolumn{4}{c}{\textbf{ResNet-18 (CIFAR-100)}} \\
\hline
\textbf{Precision} &\textbf{Acc.} (\%) & \textbf{Constraint} & \textbf{Cycles} (Ratio) \\
\hline
8-bit     & 71.48 & -                & 25.51M (1.0$\times$)  \\
Mixed     & 71.38 & 0.6$\times$BOPs  & 18.46M (\textbf{\textcolor{BrickRed}{0.72$\times$}}) \\
Mixed     & 71.31 & 0.6$\times$\textit{Proxy} & \textbf{14.75M} (\textbf{\textcolor{ForestGreen}{0.58$\times$}}) \\
\hline
\end{tabular}
\end{table}



\textcolor{revise}{As shown in Tab.~\ref{tab:vgg_on_bf}, under a 60\% constraint, the traditional BOPs-based exploration reduces the actual hardware latency to 72\%, whereas the HAP-based exploration pushes it below 60\% with negligible accuracy drop, demonstrating that hardware-aware proxies better satisfy the latency constraints.}


To analyze the precision assignments for the above experiment results, we plot the layer-by-layer bitwidths for both networks in Fig.~\ref{fig:vgg-bitwidth} and Fig.~\ref{fig:resnet18-bitwidth}. The resulting bitwidth distributions reveal key architectural insights: for example, in VGG7, two convolution layers~(2 and 4) are assigned the lowest activation bitwidths, while two linear layers~(5 and 6) are assigned the lowest weight bitwidths. 

\input{Content/Result_On_CPUs}

\section{Conclusion}
\color{revise}

We propose \textbf{MiCo}~\cite{mico_iccad}, an end-to-end mixed-precision quantization (MPQ) exploration and deployment pipeline. By combining an ensemble-model accuracy predictor with hardware-aware near-constraint sampling, MiCo efficiently identifies layer-wise quantization schemes that achieve high accuracy under hardware constraints. We further extend MiCo to \textbf{MiCoPro}, which introduces a Hardware-Aware Proxy~(HAP) model featuring network-dependent min-max calibration and inter-hardware transfer learning, drastically reducing the profiling overhead for new hardware targets.

Supported by Python APIs and specialized software kernels, the end-to-end deployment path converts the optimized MPQ models into tangible speedups on platforms ranging from SIMD-extended RISC-V CPUs to dedicated accelerators, achieving up to 40\% latency reduction with less than 3\% accuracy drop.
\color{black}




\color{revise}
\section*{Acknowledgement}
The authors used ChatGPT and DeepSeek solely for grammar editing and readability improvement in the manuscript. All suggested text was reviewed and edited by the authors, who take full responsibility for the content.
\color{black}

\bibliography{refs}
\bibliographystyle{ieeetr}

\begin{IEEEbiography}[{\includegraphics[width=1in,height=1.25in,clip,keepaspectratio]{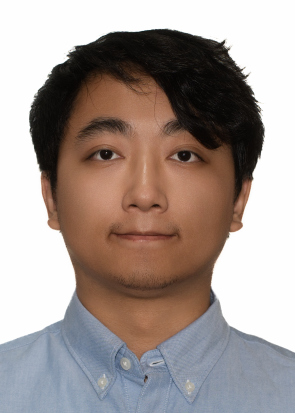}}]{Zijun Jiang}
(Student Member, IEEE) received the B.Sc. degree in Electronic Information Science from Central South University, Changsha, China, in 2022. He is currently pursuing the Ph.D. with the Microelectronics Thrust at The Hong Kong University of Science \& Technology (Guangzhou), China.

His research interests include hardware/software co-design, mixed precision neural networks, and edge AI acceleration.
\end{IEEEbiography}

\begin{IEEEbiography}[{\includegraphics[width=1in,height=1.25in,clip,keepaspectratio]{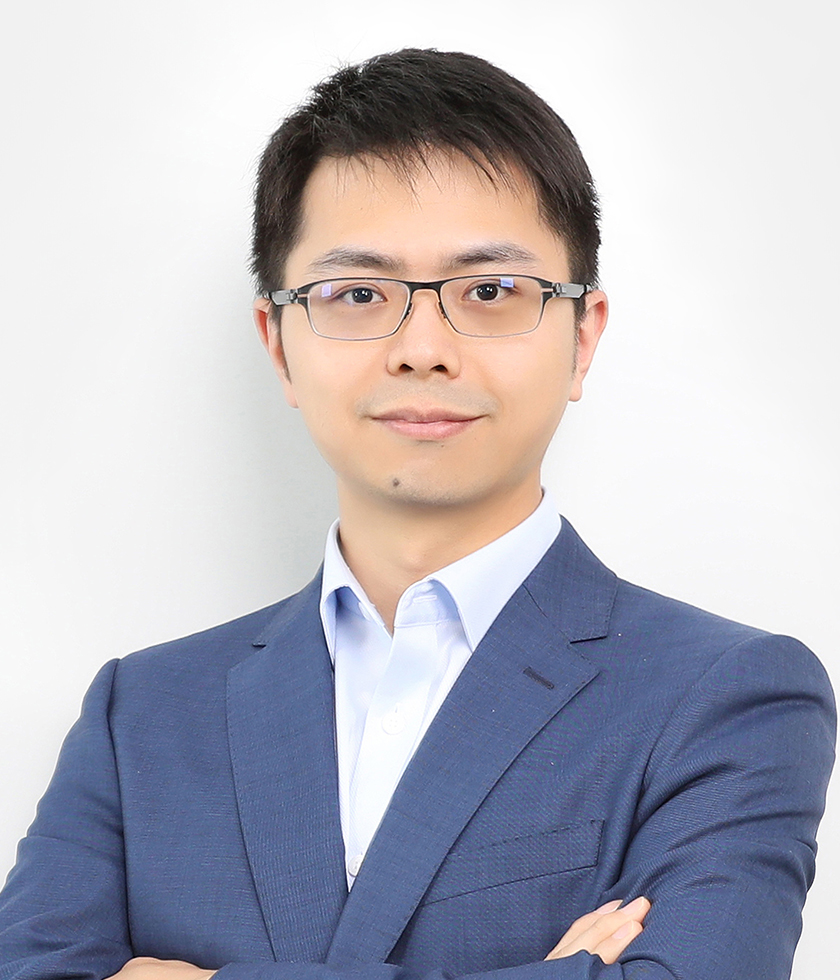}}]{Yangdi Lyu}
(Member, IEEE) received the Ph.D. degree from the Department of Computer and Information Sciences and Engineering, University of Florida, Gainesville, FL, USA, in 2020. He is an Assistant Professor with the Microelectronics Thrust, The Hong Kong University of Science and Technology (Guangzhou), Guangzhou.
His research interests include hardware security and trust, Electronic Design Automation (EDA), and test generation.
\end{IEEEbiography}

\end{document}

%% file: Content/Introduction.tex
\section{Introduction}

\IEEEPARstart{T}{iny} machine learning (ML) and edge artificial intelligence (AI) are becoming increasingly important and valuable in today's AI ecosystem. However, deploying AI models on edge devices is challenging due to the tight resource constraints. To address this, quantization techniques that convert high-precision floating-point data to low-precision integer data are widely adopted to reduce memory costs and improve the performance of edge AI models. For many AI applications, the quantization of neural networks can even reduce the data bitwidth to an extremely low range of below 4 bits, while still maintaining acceptable accuracy.

Instead of using a uniform bitwidth for all layers, layer-wise mixed
precision quantization~(MPQ) assigns different precisions to each model
layer to achieve better accuracy and efficiency. To retain the accuracy
while accelerating the inference of the quantized model, it is essential
to design MPQ schemes by considering layer sensitivity and computational
cost. As the number of layers keeps increasing in modern AI models,
manually analyzing and assigning MPQ schemes becomes impractical.
Existing search algorithms based on heuristic optimization and machine
learning techniques have been developed to explore MPQ schemes.
However, the high dimensionality and complex inter- and intra-layer
correlations of MPQ search spaces make efficient exploration
challenging. Furthermore, as the bitwidths become smaller, performing
quantization-aware training~(QAT) becomes necessary, while its
time-consuming nature restricts the number of schemes that can be
evaluated within a limited search budget.


Besides the efficient exploration of MPQ schemes, deploying MPQ models
on specific hardware platforms introduces another challenge. Existing
mixed-precision hardware designs have explored various architectures
and execution mechanisms, ranging from dedicated accelerators to
processor extensions. Application-specific accelerators provide high
performance by supporting flexible precision computation, but they
usually introduce additional hardware complexity and limited
portability. For edge devices with strict area and energy constraints,
extending general-purpose processors with mixed-precision instructions
and optimized kernels provides a more practical solution. Recent works
have demonstrated the effectiveness of RISC-V based extensions and
SIMD-style execution for accelerating low-bitwidth neural networks.


\color{revise}
However, supporting MPQ inference requires coordination among multiple
layers of the deployment stack, including quantization scheme
selection, operator implementation, data packing, and runtime support.
Existing deployment frameworks mainly target fixed-precision inference,
while hardware-specific solutions are often optimized for individual
architectures. Therefore, a unified framework that can efficiently
explore MPQ schemes according to hardware characteristics and deploy
the resulting models across different mixed-precision platforms remains
an open challenge.
\color{black}


To address these challenges in the MPQ field, in this work, we aim to develop an open-source framework\footnote{github.com/HKUSTGZ-MICS-LYU/MiCo-python} to explore optimal MPQ schemes for given AI models with a better awareness of the underlying hardware, within the limitations of search budgets.

The main contributions of this work are as follows:
\begin{itemize}
    \item A novel exploration algorithm for MPQ schemes for neural networks under certain constraints. By combining ensemble learning models (random forests) and specially designed sampling strategies, our approach enables efficient exploration of MPQ schemes, leading to lower accuracy drops under the same constraints. 
    \item A hardware-aware latency proxy modeling method that can adapt to different hardware targets. The proposed proxy model produces latency estimates that more accurately reflect the performance of real hardware compared to the traditional \textcolor{revise}{bit operations~(BOPs)} metric, effectively guiding the MPQ exploration towards the schemes that deliver true acceleration.
    \item A holistic framework to explore MPQ models and deploy them onto various hardware platforms, including accelerators and CPUs with mixed-precision acceleration support. We illustrate this framework through two case studies on different hardware platforms, where speedup is achieved by the MPQ models with minimal accuracy drops.
\end{itemize}

%% file: Content/Background.tex
\section{Background}

\subsection{Quantization}
\label{sec:MPQ}
Quantization maps high-precision continuous values (weights and activations) into a finite, discrete set of levels defined by a precision constraint. In hardware-oriented designs, symmetric uniform quantization is widely adopted because it maps zero directly to zero without requiring a zero-point offset, simplifying the arithmetic logic.

\subsubsection{General Integer Quantization ($\geq$2 bits)}
For a floating-point value $x$, the $b$-bit signed integer representation $x_q$ is expressed as:

\color{revise}

\begin{equation}
x_q = \text{clamp}\left( \left\lfloor \frac{x}{S} \right\rceil, q_{\min}, q_{\max} \right)
\end{equation}
where $\lfloor \cdot \rceil$ denotes the round-to-nearest integer operator, $\text{clamp}(v, a, b) = \max(a, \min(v, b))$, and $q_{\min}, q_{\max}$ define the integer range. For a standard $b$-bit signed integer representation, $q_{\min} = -2^{b-1}$ and $q_{\max} = 2^{b-1}-1$. Here, the scale factor $S$ represents the quantization step size between adjacent discrete levels, determined by the dynamic range of $x$:

\begin{equation}
S = \frac{\max(|x|)}{q_{\max}}
\end{equation}

\subsubsection{Extremely Low-Bit Quantization (Ternary \& Binary)}
When precision drops below 2 bits, quantization is more naturally formulated via thresholding.

\begin{itemize}
    \item Ternary Quantization. Values are constrained to $\{-1, 0, +1\}$. Its three quantization levels yield an effective bitwidth of $\log_2 3 \approx 1.58$ bits. $S$ is set to be the mean absolute value of $x$. Consequently, the general rounding and clamping operation simplifies as follows: 
    \begin{equation}
x_q =
\begin{cases}
+1, & x \geq S/2 \\
-1, & x \leq -S/2 \\
0, & otherwise\\
\end{cases}
\end{equation}
    \item Binary Quantization. Values map strictly to $\{-1, +1\}$ based on sign. 
    \begin{equation}
x_q = \text{sign}(x) =
\begin{cases}
+1, & x > 0 \\
-1, & x \le 0
\end{cases}
\end{equation}
\end{itemize}

\color{revise}
This extreme quantization is mainly applied to weights, while activations are retained at a relatively higher bitwidth to preserve the network's representational capacity~\cite{bitnet}. However, previous works such as~\cite{xnor-net, xor-net} have also applied binarization to activations for networks targeting image classification.


\color{black}

\subsubsection{Mixed Precision Quantization (MPQ)}
\color{revise}
While conventional quantization assigns a uniform bitwidth (such as INT8 or INT4) across all layers, individual network layers exhibit varying robustness to the numerical perturbations, leading to either excessive accuracy degradation or unexplored efficiency gains. \color{black}

Therefore, layer-wise mixed precision quantization, which assigns different precisions to individual layers, is introduced to improve the flexibility and efficiency of QNNs. Each layer's configuration is denoted as W$x$A$y$, where $x$ and $y$ are the weight and activation bitwidths~($b_w$, $b_a$); e.g., W4A8 indicates 4-bit weights and 8-bit activations.

\subsection{PTQ v.s. QAT}
\label{sec:ptq_vs_qat}
As the number of layers in a network increases, the search space for MPQ schemes grows exponentially. Searching this vast space requires evaluating lots of candidate layer configurations, highlighting the critical trade-off between Post-Training Quantization (PTQ) and Quantization-Aware Training (QAT).
PTQ directly quantizes a pre-trained model with the chosen schemes, being fast but often less accurate; QAT re-trains the quantized model, costing more computation and time but typically yielding higher accuracy.


Prior studies on low-precision sensitivity~\cite{zhao2023_ptq_qat} demonstrate that low-bitwidth schemes (e.g., INT4 and INT2) suffer from severe quantization noise under PTQ. To establish the precise transition point where QAT becomes necessary in our setup,
we evaluate the QAT–PTQ accuracy gap on \textcolor{revise}{a convolutional neural network (CNN) model from \cite{cmsis-cnn} and ResNet-18~\cite{resnet} trained on the CIFAR-10/CIFAR-100 dataset when its second layer} is quantized to different weight and activation precisions.

\input{PTQ_vs_QAT}
As shown in Fig.~\ref{fig:ptq_vs_qat}, the accuracy difference between QAT and PTQ is small for bitwidths of 4 bits or higher~(e.g., W4A8 and W4A4), but becomes significant for bitwidths below 4. This result motivates a bifurcated MPQ evaluation strategy: leveraging rapid PTQ estimation for candidates at or above 4bits, while applying QAT for sub-4-bit configurations to maintain high accuracy. 

\subsection{Hardware-Aware MPQ Exploration}
\label{sec:mpq_bops}
\color{revise}
To efficiently explore MPQ schemes, it is essential to estimate the
hardware cost of candidate models. Direct latency evaluation relies on
cycle-accurate simulation or on-device profiling, which can take seconds
to minutes depending on the target architecture and model complexity.
Consequently, analytical metrics such as Bit Operations~(BOPs) are widely
adopted to approximate the computational cost of mixed-precision models
without repeatedly invoking expensive hardware evaluations~\cite{uniq}.

The BOPs of the $i$-th layer are defined as
\begin{equation}
    \mathrm{BOPs}_i =
    b_{w_i} b_{a_i} \mathrm{MACs}_i,
    \label{eq:bops}
\end{equation}
where $b_{w_i}$ and $b_{a_i}$ denote the weight and activation
bitwidths, respectively, and $\mathrm{MACs}_i$ is the number of MAC
operations in the $i$-th layer.

More generally, hardware-aware MPQ exploration can be formulated as
\begin{equation}
    \begin{aligned}
        \text{Maximize: }
        & \mathrm{Accuracy}(M_Q), \\
        \text{Subject to: }
        & C_{\mathcal{H}}(Q) \leq C_{\mathrm{constr}},
    \end{aligned}
    \label{eq:opt}
\end{equation}
where $M_Q$ is the model quantized using scheme
$Q=[(b_{w_1},b_{a_1}),\ldots,(b_{w_L},b_{a_L})]$, and
$C_{\mathcal{H}}(Q)$ represents its hardware cost on target
$\mathcal{H}$, which can be instantiated using analytical metrics or
target-specific performance estimates. In previous MPQ
frameworks~\cite{hawqV3,bayesian_bits,edge-mpq}, it is commonly defined
as the total BOPs, $\sum_{i=1}^{L}\mathrm{BOPs}_i$, due to its low
evaluation overhead.
\color{black}


Although the BOPs serves as an efficient hardware-cost proxy, it primarily captures theoretical arithmetic complexity without capturing how quantized operators execute on target hardware. \textcolor{revise}{In practice, real-world execution gains are influenced by hardware features such as mixed-precision processing width and parallelism, SIMD/vector lane utilization, operand packing into registers, memory access and bus widths, and data packing/unpacking overhead. These factors determine whether a reduction in bitwidth can translate into higher effective parallelism and lower memory traffic. Consequently, identical BOP reductions can lead to substantially different end-to-end latency improvements across different hardware targets.}

\begin{figure}[b]
    \centering
    \begin{subfigure}{0.22\textwidth}
        \includegraphics[width=\textwidth]
        {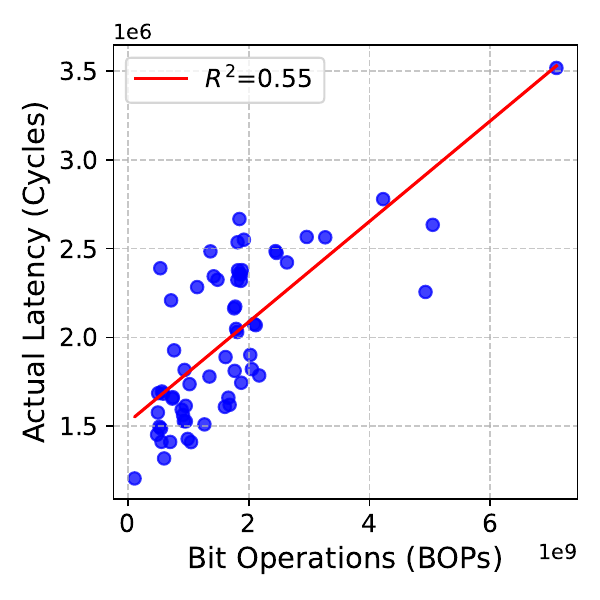}
        \caption{VGG7 on BitFusion}
        \label{fig:latency_on_accelerator}
    \end{subfigure}%
    \begin{subfigure}{0.22\textwidth}
        \includegraphics[width=\textwidth]
        {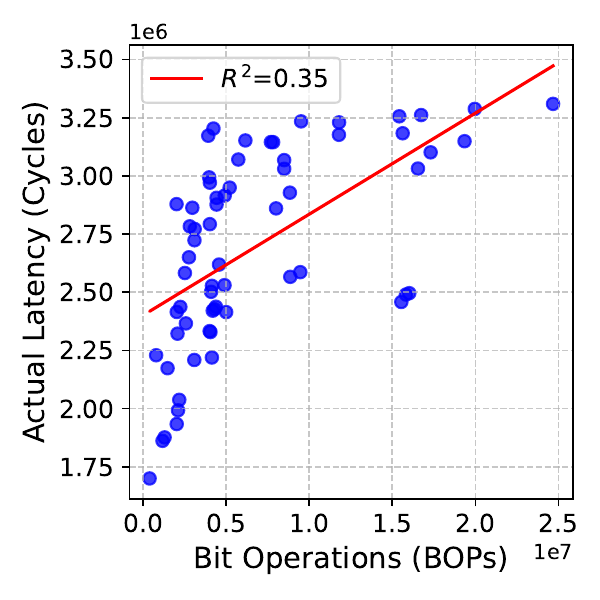}
        \caption{LeNet on Extended CPU}
        \label{fig:latency_on_cpu}
    \end{subfigure}
    \caption{Correlation between BOPs and actual latency.}
    \label{fig:corr_bops}
\end{figure}

We randomly sampled 64 MPQ schemes for VGG7~\cite{vgg} and
LeNet~\cite{lenet} and measured their actual latencies on
BitFusion~\cite{bit_fusion} and VexiiMiCo~\cite{mico_iccad}. As shown
in Fig.~\ref{fig:corr_bops}, BOPs is positively correlated with latency,
but the low $R^2$ values show it does not fully capture target-specific
end-to-end performance, motivating the Hardware-Aware Proxy~(HAP) model
introduced later.

%% file: PTQ_vs_QAT.tex
\definecolor{CNNQAT}{HTML}{94FFD8}
\definecolor{CNNPTQ}{HTML}{A3D8FF}
\definecolor{ResQAT}{HTML}{FFD580}
\definecolor{ResPTQ}{HTML}{F7A6A6}

\begin{figure}[b]
\centering
\scalebox{0.9}{%
\begin{tikzpicture}
\begin{axis}[
    ybar,
    height=1.2in,
    width=0.4\textwidth,
    scale only axis,
    ylabel={Top-1 Accuracy},
    ymin=0.2,
    ymax=0.85,
    xmin=0.5,
    xmax=6.5,
    xtick={1,2,3,4,5,6},
    xticklabels={
        W4A8,
        W4A4,
        W2A4,
        W2A2,
        W1A2,
        W1A1
    },
    bar width=8pt,
    axis on top,
    legend style={
        at={(0.5,-0.25)},
        anchor=north,
        legend columns=2,
        draw=none,
        font=\small,
        /tikz/every even column/.append style={
            column sep=4pt
        }
    },
    y label style={
        at={(-0.1,0.5)},
        anchor=south
    }
]


\addplot[
    fill=CNNQAT,
    draw=black,
    bar shift=-6pt,
    forget plot
] coordinates {
    (1,0.7815)
    (2,0.7776)
    (3,0.7670)
    (4,0.7108)
    (5,0.6839)
    (6,0.6290)
};

\addplot[
    fill=CNNPTQ,
    draw=black,
    bar shift=-6pt,
    forget plot
] coordinates {
    (1,0.7129)
    (2,0.7039)
    (3,0.3111)
    (4,0.2697)
    (5,0.3347)
    (6,0.2935)
};


\addplot[
    fill=ResQAT,
    draw=black,
    bar shift=6pt,
    forget plot
] coordinates {
    (1,0.7086)
    (2,0.7062)
    (3,0.7047)
    (4,0.6963)
    (5,0.6948)
    (6,0.6931)
};

\addplot[
    fill=ResPTQ,
    draw=black,
    bar shift=6pt,
    forget plot
] coordinates {
    (1,0.7060)
    (2,0.7039)
    (3,0.6851)
    (4,0.5543)
    (5,0.2579)
    (6,0.3254)
};


\addlegendimage{
    area legend,
    fill=CNNPTQ,
    draw=black
}
\addlegendentry{CNN PTQ}

\addlegendimage{
    area legend,
    fill=CNNQAT,
    draw=black
}
\addlegendentry{CNN QAT Recovery}

\addlegendimage{
    area legend,
    fill=ResPTQ,
    draw=black
}
\addlegendentry{ResNet-18 PTQ}

\addlegendimage{
    area legend,
    fill=ResQAT,
    draw=black
}
\addlegendentry{ResNet-18 QAT Recovery}

\end{axis}
\end{tikzpicture}
}
\caption{PTQ Accuracy and QAT Recovery}
\label{fig:ptq_vs_qat}
\end{figure}

%% file: Content/RelatedWork.tex
\color{revise}

\section{Related Work}
\label{sec:related_work}

\subsection{Mixed Precision Quantization Exploration}

Existing MPQ exploration methods can be broadly categorized
into analytical optimization, learning-based search, and differentiable
optimization approaches. A comparison of representative MPQ exploration
methods is summarized in Tab.~\ref{tab:SW_Cmp}.

\begin{table}[ht]
    \centering
    \color{revise}
    \caption{Comparison with Existing Algorithms}
    \label{tab:SW_Cmp}
    \begin{threeparttable}
    \begin{tabular}{ l | c l c c}
    \hline
    \textbf{Work}  & \textbf{Method} & \textbf{Bitwidths}& Support $b_a \neq b_w$? & $B$ \\
    \hline
    HAWQ-V3~\cite{hawqV3}     & ILP & 8/4        & \xmark                & 2    \\
    $w$-based~\cite{edge-mpq} & NLP & 8/7/6/5/4  & \xmark                & 5    \\
    BOMP~\cite{van2023bomp}   & BO  & 8/7/6/5/4  & \xmark                & 5    \\
    HAQ~\cite{haq}            & RL  & 8/6/4/2    & \color{teal}{\cmark}  & 16   \\
    Bayes. Bits~\cite{bayesian_bits} & GB & 16/8/4/2 & \color{teal}{\cmark} & 16 \\
    Differ.~\cite{schaefer2024differentiable} & GB & 8/7/6/5/4/3/2 & \color{teal}{\cmark} & 49 \\ 
    \hline
    \textbf{MiCo}~(PTQ)       & RF+NCS  & 8/7/6/5/4   & \color{teal}{\cmark} & \textbf{25}     \\
    \textbf{MiCo}~(QAT)       & RF+NCS  & 8/4/2/1 & \color{teal}{\cmark} & \textbf{16}     \\
    \hline
    \end{tabular}
\begin{tablenotes}
\item \textcolor{revise}{$B$: Number of Possible Weight/Activation Bitwidth Combinations}
\end{tablenotes}
\end{threeparttable}
\end{table}

Analytical optimization methods estimate layer-wise quantization
sensitivity and formulate MPQ assignment as a constrained optimization
problem. HAWQ-V3~\cite{hawqV3} employs Hessian-aware sensitivity
analysis with integer linear programming (ILP) to determine the
bitwidth assignment under hardware constraints. The $w$-based
method~\cite{edge-mpq} models the impact of quantization on each layer
and solves the resulting MPQ assignment problem using nonlinear
programming (NLP). These approaches efficiently reduce the search cost
by exploiting layer-wise statistics. However, they generally rely on
predefined sensitivity metrics and provide limited flexibility in
exploring independent weight and activation precisions.

Learning-based methods treat MPQ exploration as a black-box
optimization problem and learn the relationship between quantization
schemes and model accuracy. HAQ~\cite{haq} adopts reinforcement learning
to sequentially determine layer-wise bitwidths, enabling flexible
weight and activation precision selection. BOMP~\cite{van2023bomp}
utilizes Bayesian optimization with Gaussian process models to improve
sample efficiency compared with exhaustive search. Nevertheless,
reinforcement-learning-based methods typically require a large number
of evaluations to train the policy, while Gaussian-process-based
optimization may suffer from scalability issues when the search
dimension increases with the number of network layers.

Another research direction incorporates precision selection into the
training process through gradient-based~(GB) differentiable
optimization. Bayesian Bits~\cite{bayesian_bits} introduces learnable
bitwidth variables with stochastic gates, enabling gradient-based
optimization of mixed-precision configurations. Schaefer \emph{et al.}
\cite{schaefer2024differentiable} further propose a fully differentiable
quantization framework with hardware-aware objectives for edge
inference. These methods jointly optimize network parameters and
quantization choices, and can achieve effective accuracy--efficiency
trade-offs. However, the additional learnable precision variables,
gating parameters, and differentiable quantization operators enlarge
the optimization state and introduce extra forward- and backward-pass
overhead. As a result, they generally require a dedicated end-to-end
training procedure and can noticeably slow down QAT, especially for
large models or wide mixed-precision search spaces. This makes them less
suitable for black-box exploration under strict evaluation and training
budgets.

MiCo~\cite{mico_iccad} adopts the RF-guided near-constraint
sampling~(RF+NCS) strategy detailed in
Sec.~\ref{sec:exploration_flow} for efficient exploration under
limited PTQ/QAT evaluation budgets. MiCoPro further extends the
framework with hardware-aware latency modeling and multi-platform
deployment support.

\subsection{Mixed Precision Hardware and Deployment}

A representative selection of mixed-precision inference hardware architectures is summarized in Tab.~\ref{tab:MP_Hardware}. The reported
performance values are directly adopted from the corresponding
publications for reference, as different works may use different
technology nodes, frequencies, and evaluation configurations.
Dedicated accelerators such as BitFusion~\cite{bit_fusion} employ configurable
low-bitwidth datapaths to provide high computational throughput.
SPEED~\cite{speed}, in contrast, is a scalable RISC-V vector processor
that extends the RISC-V Vector ISA with customized multi-precision
instructions. It integrates a parameterized multi-precision tensor unit
and flexible dataflow mapping to support DNN operators with precisions
ranging from 4 to 16 bits.

\begin{table}[ht]
    \centering
    \color{revise}
    \begin{threeparttable}
    \caption{State-of-the-art Mixed Precision Hardware}
    \label{tab:MP_Hardware}
    \begin{tabular}{ c | c | l | l }
    \hline
    \textbf{Type} & \textbf{Hardware} & \textbf{Bitwidths} & \textbf{Perf.(GOPs)} \\
    \hline
    \multirow{2}{*}{\textbf{Accelerator}} & BitFusion~\cite{bit_fusion} & 8b-2b &  - \\
    & SPEED~\cite{speed} & 16b/8b/4b & 343.1-737.9 \\
    \hline
    \multirow{7}{*}{\textbf{CPU Extension}} & XpulpNN~\cite{xpulpnn} & 16/8b/4b/2b   & 19.0-48.0 \\
    & MPIC~\cite{MPIC}             & 8b/4b/2b      & 1.1-3.3 \\
    & Mixed-GEMM~\cite{mix-gemm}   & 8b-2b         & 4.2-7.9   \\
    & MPQ VMM~\cite{edge-mpq}      & 16b/8b/4b/2b  & 0.93-2.86 \\
    & Multi-Pump~\cite{multi-pump} & 8b/4b/2b      & 0.24-0.85 \\
    & Customized~\cite{flexible-mp}           & 8b-2b         & 3.3 \\
    & STAR-MAC~\cite{star-mac}     & 16b/8b/4b     & 0.5-2.0 \\
    \hline
    \end{tabular}
    \end{threeparttable}
\end{table}

For resource-constrained edge systems, many studies extend processors
with packed-SIMD instructions, configurable arithmetic units, and
optimized low-bitwidth kernels. XpulpNN~\cite{xpulpnn},
MPIC~\cite{MPIC}, Mixed-GEMM~\cite{mix-gemm}, and MPQ
VMM~\cite{edge-mpq} improve mixed-precision execution through
instruction-set and software-kernel extensions. Armeniakos \emph{et al.}
\cite{multi-pump} propose multi-pumped soft-SIMD operations together
with a cross-layer hardware--software flow. STAR-MAC~\cite{star-mac}
integrates a reconfigurable mixed-precision MAC unit into an Ibex core
and provides quantization and mixed-precision deployment support. Lu \emph{et al.}
\cite{flexible-mp} similarly present an integrated software--hardware
co-design framework with customized RISC-V instructions, data packing,
and optimized inference kernels.

General deployment frameworks such as TVM~\cite{TVM} and
DORY~\cite{dory} provide compilation and runtime support for edge
inference, but their standard flows mainly target fixed-precision
operators. Supporting MPQ models therefore requires coordination among
quantization, data packing, operator lowering, and target-specific
kernels. Existing hardware--software co-design solutions provide such
support for their respective architectures, whereas MiCoPro focuses on
adapting MPQ exploration across different hardware targets. It combines
a reusable target-specific latency proxy with MPQ search and deployment,
rather than optimizing only a single processor architecture or relying
on a hardware-independent metric such as bit operations.

\color{black}

%% file: Content/MPQ_Search.tex
\section{MiCo: Mixed-Precision Quantization Exploration}
\label{sec:exploration_flow}
\color{revise}
We formulate MPQ exploration as a budget-constrained black-box
optimization problem and adopt an RF-guided near-constraint sampling
strategy, denoted as RF+NCS. The RF surrogate predicts the accuracies
of complete MPQ schemes, while layer-wise orthogonal sampling constructs
an informative initial training set. The hardware cost can be defined
using BOPs, model size, or target-specific end-to-end latency estimated
by the Hardware-Aware Proxy~(HAP) introduced in
Sec.~\ref{sec:hap}. NCS then generates feasible candidates close to the
specified constraint, which are ranked by the RF surrogate and evaluated
through PTQ or short-QAT.

\begin{figure}[b]
    \centering
    \includegraphics[width=0.45\textwidth]{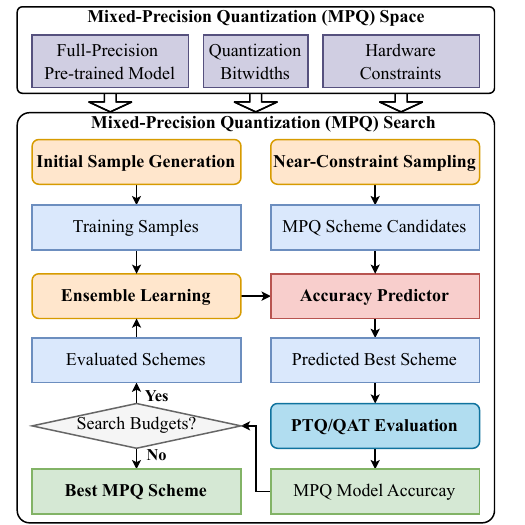}
    \caption{Overview of the MiCo MPQ exploration flow.}
    \label{fig:search_flow}
\end{figure}

The complete exploration procedure is illustrated in
Fig.~\ref{fig:search_flow} and summarized in
Alg.~\ref{alg:mpq_search}. After evaluating the initial schemes,
MiCo iteratively trains the RF surrogate, generates near-constraint
candidates, and evaluates the highest-ranked unevaluated scheme. Each
new observation is added to the training set to update the surrogate.
Once the evaluation budget is exhausted, the feasible evaluated scheme
with the highest accuracy is selected for deployment.
\color{black}

\begin{algorithm}[htbp]
\caption{MPQ Search Loop}
\label{alg:mpq_search}
\color{revise}

\KwIn{
$N_{\mathrm{init}}$--number of initial samples,
$N_s$--number of search iterations,
$N_c$--number of candidate schemes,
$M$--MPQ model,
$C_{\mathcal{H}}(\cdot)$--hardware-cost function,
$C_{\mathrm{constr}}$--constraint,
$\rho$--ROI width
}
\KwOut{$Q_{\mathrm{best}}$--best MPQ scheme}

$\mathcal{S}_{\mathrm{train}}
\gets
\textit{InitialSampling}(N_{\mathrm{init}},M)$\;

\For{$t\gets1$ \KwTo $N_s$}{
    $f_t\gets\textit{TrainRF}(\mathcal{S}_{\mathrm{train}})$\;

    $\mathcal{P}_t\gets
    \textit{NearConstraintSampling}
    (N_c,C_{\mathcal{H}},C_{\mathrm{constr}},\rho)$\;

    $Q_t\gets
    \underset{\substack{Q\in\mathcal{P}_t}}
    {\arg\max}\ f_t(Q)$\;

    $Acc_t\gets$ Evaluate $M_{Q_t}$ on the test dataset\;

    $\mathcal{S}_{\mathrm{train}}
    \gets
    \mathcal{S}_{\mathrm{train}}
    \cup\{(Q_t,Acc_t)\}$\;
}

$Q_{\mathrm{best}}\gets
\underset{\substack{(Q,Acc)\in\mathcal{S}_{\mathrm{train}}\\
C_{\mathcal{H}}(Q)\leq C_{\mathrm{constr}}}}
{\arg\max}\ Acc$\;

\Return $Q_{\mathrm{best}}$\;

\color{black}
\end{algorithm}

\subsection{Ensemble Learning-based MPQ Accuracy Predictor}
\color{revise}
Evaluating an MPQ scheme $Q$ on the test set is computationally expensive. We therefore train an accuracy predictor using a small set of sampled MPQ schemes and their measured accuracies, and use it as a low-cost proxy during the search.

We adopt Random Forest~(RF)~\cite{random_forest} because it can efficiently model complex non-linear interactions among discrete bitwidth assignments while remaining robust with limited training samples. Compared with Gaussian Processes, RF scales better to high-dimensional search spaces and avoids kernel sensitivity; compared with RL-based methods, it requires substantially fewer costly accuracy evaluations. Its effectiveness as a predictor has also been demonstrated in recent NAS studies~\cite{NAS_weak, gp_nas_ensemble}. To further improve prediction stability, we augment each layer's bitwidth feature with its number of MACs, allowing the predictor to account for differences in layer-wise computational significance.
\color{black}

\subsection{Layer-wise Orthogonal Initial Sampling}
\label{sec:initial}

For learning-based algorithms, the representative initial samples can strengthen the performance of predictive models like random forest models and Gaussian process models. A common way of generating initial samples is through uniform random sampling of the search space. However, we have found that random sampling is very inefficient for MPQ problems due to their extremely large space.

To address this issue, we propose a layer-wise orthogonal initial sampling method, which is described in Alg.~\ref{alg:init}. First, to make sure the algorithm learns the upper and lower bounds of accuracy, we include the MPQ schemes with the highest and lowest bitwidths~(e.g., W8A8 and W1A1) in the initial samples. 
Then, for each subsequent initial sample, the method randomly selects several layers and quantizes them with random schemes, while keeping other layers in the W8A8 scheme. 
\color{revise2}
The sampling procedure constructs a layer-wise orthogonal subset through non-overlapping layer selections until all layers are covered. If any initial-sample budget remains, previously covered layers are eligible for re-selection, ensuring valid indexing while guaranteeing that every layer is quantized at least once.
\color{black}
\begin{algorithm}[t]
\caption{Layer-wise Orthogonal Initial Sampling}\label{alg:init}
    \KwIn{$N_{init}$-Number of Initial Samples, $M$-MPQ Model}
    \KwOut{$\mathcal{S}$-Initial Samples}
    $\mathcal{S} \gets \emptyset$; \\
    $b_{\max} \gets$ Largest supported bitwidth; \\
    $b_{\min} \gets$ Lowest supported bitwidth; \\
    $Q_1 \gets [(b_{\max},b_{\max}),\cdots,(b_{\max}, b_{\max})]$; \\
    $Q_2 \gets [(b_{\min},b_{\min}),\cdots,(b_{\min}, b_{\min})]$; \\
    Randomly permute the layer indices and get $[l_1, l_2, \cdots, l_L]$;\\
    \color{revise} $N_m \gets \left\lceil L/(N_{\mathrm{init}}-2)\right\rceil$ \color{black} \\
    \For{$i=3 \to N_{init}$}{
        $Q_i \gets [(b_{\max},b_{\max}),\cdots,(b_{\max}, b_{\max})]$; \\
        Modify the layers starting from $start = (i - 3) * N_m$; \\
        \For{$k=1 \to N_{m}$}{
            \color{revise}
            $p \gets start + k$; \\
            \eIf{$p \leq L$}{
                $l \gets l_p$; \\
            }{
                Randomly select $l\in\{1,\ldots,L\}$; \\
            }
            Randomly modify the weight and activation bitwidths of layer $Q_i[l]$\;
            \color{black}
        }
    }
    \For{$i=1 \to N_{init}$}{
        $Acc_i \gets$ Evaluate $M_{Q_i}$ on the test dataset; \\
        \color{revise} $\mathcal{S} \gets \mathcal{S} \cup \{(Q_i, Acc_i)\}$; \color{black}
    }
    \Return $\mathcal{S}$;
\end{algorithm}

With the above sampling method, the effects of different quantized layers and quantization schemes on accuracy are effectively captured in the initial samples, enabling our prediction model to acquire valuable insights.
\color{revise}
\subsection{Near-Constraint Sampling (NCS) and Optimization}
\label{sec:optimization}
\color{black}

\begin{algorithm}[htbp]
\caption{Near-Constraint Sampling}
\label{alg:ncs}
\color{revise}
\KwIn{
$N$--number of samples,
$C_{\mathcal{H}}(\cdot)$--hardware-cost function,
$C_{\mathrm{constr}}$--constraint,
$\rho$--ROI width
}
\KwOut{$\mathcal{P}$--near-constraint schemes}

$b_{\max} \gets$ Largest supported bitwidth; \\
$Q_{\max} \gets [(b_{\max},b_{\max}),\cdots,(b_{\max}, b_{\max})]$; \\
$C_{\max}\gets C_{\mathcal{H}}(Q_{\max})$\;
$\mathcal{R}\gets
[C_{\mathrm{constr}}-\rho C_{\max},\,
 C_{\mathrm{constr}}]$\;

Define $d(Q)$ as the distance from $C_{\mathcal{H}}(Q)$ to $\mathcal{R}$\;
$\mathcal{P}\gets$ $N$ random uniform quantization schemes\;

\For{$g\gets1$ \KwTo $G_{\max}$}{
    \While{$|\mathcal{P}|<2N$}{
        Randomly select two parents $Q_a,Q_b\in\mathcal{P}$\; 
        Generate $Q$ from $Q_a$ and $Q_b$ by element-wise
        crossover and mutation\;
        \If{$Q\notin\mathcal{P}$}{
            $\mathcal{P}\gets\mathcal{P}\cup\{Q\}$\;
        }
    }

    Sort $\mathcal{P}$ in ascending order of $d(Q)$\;
    Retain the first $N$ schemes in $\mathcal{P}$\;

    \If{$\forall Q\in\mathcal{P}: d(Q)=0$}{
        \textbf{break}\;
    }
}

\Return $\mathcal{P}$\;
\color{black}
\end{algorithm}

\begin{figure}[htbp]
    \centering
    \includegraphics[width=0.45\textwidth]{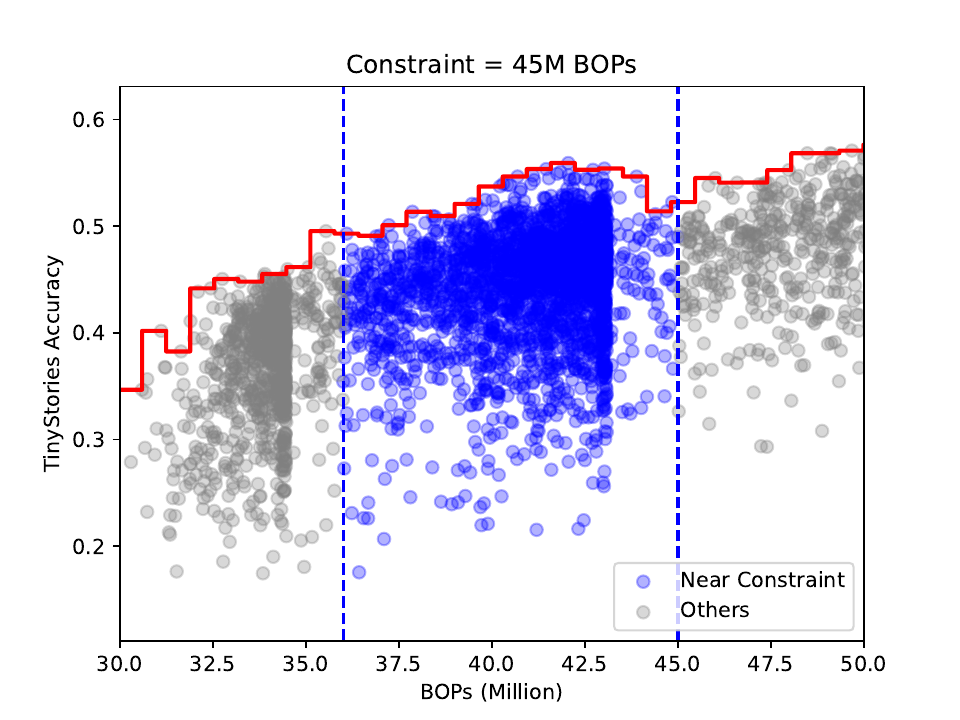}
    \caption{Accuracy v.s. BOPs for TinyLLaMa on TinyStories}
    \label{fig:acc_vs_bops}
\end{figure}

For a given hardware cost $x$, we define the accuracy upper envelope as
$F(x)=\max_{Q:C(Q)=x}\mathrm{Accuracy}(M_Q)$. The objective in
Eq.~\ref{eq:opt} is therefore to maximize $F(x)$ subject to
$x\leq C_{\mathrm{constr}}$. Although local fluctuations exist, $F(x)$
generally increases with the available hardware budget because
higher-cost schemes tend to preserve more layers at high precision.
This trend is also observed from randomly sampled TinyLLaMa schemes in
Fig.~\ref{fig:acc_vs_bops}, where each point denotes an evaluated
scheme and the red curve represents the observed accuracy envelope.

This motivates us to focus the search space near the constraint. As described in Alg.~\ref{alg:ncs}, we integrate a standard genetic algorithm to generate samples within \textcolor{revise}{a targeted region of interest (ROI)}, and evaluate the predicted accuracy within the near-constraint space. Initially, the ROI is restricted to \textcolor{revise}{$[0.8*C_{\text{constr}}, C_{\text{constr}}]$}, and it is gradually expanded to \textcolor{revise}{$[0.5*C_{\text{constr}}, C_{\text{constr}}]$} towards the end of the exploration. This strategy prioritizes schemes close to the constraints to rapidly obtain good solutions, while progressively broadening the search in case we discard promising options. 

Furthermore, we incorporate the common heuristic of maintaining high precision for the first and last layers in our sampling method, which has been proven to be effective in previous works~\cite{haq, multi-pump}.

\subsection{Accuracy Evaluation}

In each search iteration, the predicted best schemes are evaluated to get the actual accuracy of the MPQ models. In our flow, we use different evaluation processes for PTQ and QAT search. While PTQ accuracy is directly validated, the QAT accuracy is validated after a very short re-training~($\approx$ pre-trained time/20). This short re-training costs less time than a full re-training, while offering the algorithm a good approximation of the QAT recoverable accuracy mentioned in Sec.~\ref{sec:ptq_vs_qat}.

%% file: Content/Hardware_Modeling.tex
\section{MiCoPro: MiCo with Hardware-aware Proxy~(HAP) Model}
\label{sec:hap}

\begin{figure}[t]
    \centering
    \includegraphics[width=0.45\textwidth]{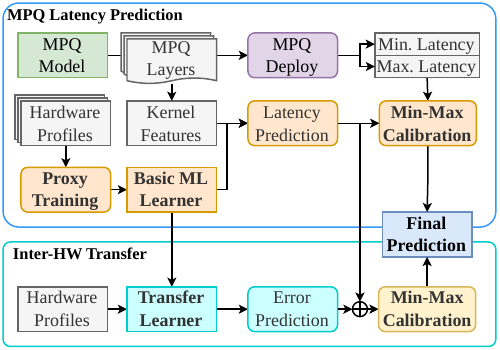}
    \caption{Overview of MiCo Hardware-aware Proxy Model}
    \label{fig:HAP}
\end{figure}

As discussed in Sec.~\ref{sec:mpq_bops}, while the BOPs metric is easy to estimate, its limitations make it less reliable for a truly hardware-aware exploration. To overcome the shortcomings of BOPs, we propose a hardware-aware proxy modeling method and integrate it into the MiCo framework for practical optimization that leads to real speedup on target hardware. The overview of the proposed modeling method is illustrated in Fig.~\ref{fig:HAP}.

\subsection{Kernel Profiling}

\label{sec:profiling}
To train hardware-aware proxy models, it is necessary to collect profile datasets for mixed-precision kernels on target hardware platforms. We collect the kernel parameters from a variety of neural networks and use them as a profiling list. The hardware profiles are collected through benchmarking the kernels with different shapes and parameters (e.g., stride and padding in convolution kernels), and also different precision combinations (e.g., W8A8, W4A8) via cycle-accurate simulations. The profiling process is a one-time effort for each hardware platform. Once collected, the dataset can be used for all MPQ exploration and deployment on the target hardware.

\subsection{Kernel Features}
\label{sec:features}
In previous works on hardware latency predictors, such as nn-Meter~\cite{nnmeter}, different sets of features are used for different kernels. For linear kernels, the feature set includes input shape, output shape, and number of operations~(FLOPs). And for convolutional kernels, the feature set is expanded with arguments like stride size and kernel size. These features are enough to build accurate predictors for fixed-precision models, but when it comes to mixed-precision models, the bitwidth-dependent features are crucial to be included. While BOPs can be such a feature, they blur the underlying hardware mechanisms, making it difficult to become hardware-aware. 

Therefore, we propose a new set of features, the \textbf{Composite Bit Operations~(CBOPs)}, to extend the features for mixed-precision kernel modeling.

CBOPs comprise three new features: Bit Multiply-Accumulate Operations (BMACs), Activation Loads (ALoads), and Weight Loads (WLoads), which are defined as follows:
\begin{equation}
\begin{split}
    &\text{BMACs}_i = \max(b_{w_i}, b_{a_i}) \text{MACs}_i \\
    &\text{ALoads}_i = b_{a_i} \text{MACs}_i  \\
    &\text{WLoads}_i = b_{w_i} \text{MACs}_i
\end{split}
\end{equation}

\color{revise}
BMACs models the total bit‑processing volume of the MAC units. The use of $\max(b_w, b_a)$ reflects a common architectural bottleneck: when operands have mixed bitwidths, the hardware typically extends the narrower operand to match the wider datapath and reuses the wider multipliers, so the larger bitwidth bounds the effective computation width and peak throughput.
The separate ALoads and WLoads terms quantify the memory traffic for activations and weights individually. This distinction captures the asymmetric impact of weight and activation bitwidths on memory bandwidth and data packing — aspects that the original BOPs product cannot differentiate. Together, these three features form a more architecture‑aware cost proxy.

\color{black}


Pure CBOPs features are sufficient to construct lightweight hardware proxies \textcolor{revise}{with simple linear regression models, offering fast estimation during MPQ exploration. For scenarios requiring the highest possible fidelity,} our framework also supports fine-grained modelling using \textbf{full kernel features}. 
This robust configuration extends the features used in nn-Meter~\cite{nnmeter} with explicit bitwidth and architectural packing features, specifically: $b_{w_i}$, $b_{a_i}$, $\lfloor 32/b_{w_i} \rfloor$, $\lfloor 32/b_{a_i} \rfloor$, $\text{MACs}_i$, and an asymmetric precision indicator ($b_{w_i} \neq b_{a_i}$). The term $\lfloor 32/b \rfloor$ explicitly models sub-word packing efficiency into standard 32-bit registers or memory buses.
When fit by machine learning-based learners~(e.g., Random Forest, XGBoosting), these full features yield superior latency prediction accuracy by capturing intricate micro-architectural behaviours. \textcolor{revise}{However, due to their more complex correlations, simple linear regression models cannot be reliably fitted with the full kernel features.}

\textcolor{revise}{Overall, CBOPs are well-suited for pairing with a simple linear model to provide fast and sufficiently accurate latency predictions, while full kernel features, combined with non-linear ML models, deliver the most accurate predictions for exploration with strict constraints.}

\subsection{Network-dependent Min-Max Calibration}

\textcolor{revise}{In practical end-to-end prediction, the network latency is not simply the sum of the latencies of computational kernels. This is because non-computational overheads and unaccelerated operations (e.g., im2col) also contribute to the end-to-end latency.} Therefore, to further enhance the accuracy and correlation of our proxy models with actual hardware latency across different networks, we apply \textbf{network-dependent min-max calibration}.

\color{revise}
Firstly, we fit a basic latency proxy model $M$, which predicts the latency of each kernel from the kernel features $x_i$ (CBOPs or full features). The resulting basic latency prediction ${Lat}_{proxy}$ is formulated as:
\begin{equation}
    {Lat}_{proxy}(\mathbf{x}) = \sum_{i=1}^L M(x_i)
\end{equation}
where $\mathbf{x} = \{x_1,x_2,\cdots,x_L\}$ containing the features for each layer. Secondly, we evaluate the minimal-precision network~(e.g., INT2) and the maximum-precision network~(e.g., INT8) on the target hardware to obtain the actual latencies of the minimal- and maximal-precision networks ($Lat_{hw}^{min},Lat_{hw}^{max}$), and also the corresponding basic predictions ($Lat_{proxy}^{min},Lat_{proxy}^{max}$). Utilizing the latency, a simple linear calibration can be applied to the basic prediction, adapting it to the target network. The calibrated proxy model $\hat{Lat}_{proxy}$ can be formulated as:
\begin{equation}
\begin{split}
& \alpha = (Lat_{hw}^{max} - Lat_{hw}^{min})/(Lat_{proxy}^{max} - Lat_{proxy}^{min}) \\
& \beta= Lat_{hw}^{min}-\alpha Lat_{proxy}^{min} \\
& \hat{Lat}_{proxy} = \alpha \cdot Lat_{proxy}(\mathbf{x}) + \beta
\end{split}
\end{equation}
The calibration method aligns the proxy predictions with the measured hardware latencies at the two precision extremes, effectively de-biasing the overhead that is unrelated to precision and thus delivering a more accurate prediction.
\color{black}

\subsection{Inter-Hardware Transfer Learning}

As discussed in Sec.~\ref{sec:profiling}, the profiling process for each hardware platform is a one-time effort. However, fine-grained profiling based on RTL simulation can be time-consuming, costing hours to days.

We find that, for mixed-precision hardware with similar acceleration architectures (e.g., different CPUs with the same SIMD extension), their latencies exhibit a similar distribution and correlation with bitwidths. This inspires us to apply the \textbf{transfer learning} technique, which largely reduces the profiling time on new MP hardware.

The lower part of Fig.~\ref{fig:HAP} illustrates the brief concept of the transfer learning proxy, which utilizes another \textbf{transfer learner} along with the basic learner that has already been trained. With only a small amount of profiling data on new hardware, the transfer learner doesn't directly fit the kernel features and corresponding latency. Instead, it learns the \textbf{residual error} of the trained learner prediction on old hardware and the actual latency on new hardware. The final proxy model can be formulated as:

\begin{equation}
    Lat_{proxy}(x) = Lat_{proxy}'(x) + \delta_{t}(x)
\end{equation}
where $Lat_{proxy}'(x)$ is the original proxy model trained on existing profile data, and $\delta_t(x)$ is the transfer learner fit on the residual error on the new profile data.

With this method, the amount of profile data required for new hardware with a similar architecture can be largely reduced, thus saving the time cost of sampling/simulating.

%% file: Content/EndToEnd_Deployment.tex
\section{End-to-End Exploration \& Deployment}
\label{sec:deployment_flow}
The MiCo framework enables the end-to-end exploration and deployment of MPQ models for various hardware targets. To achieve this, we incorporate a hardware-aware proxy model into the exploration flow explained in Sec.~\ref{sec:exploration_flow} to improve the estimation of the real end-to-end latency. 

\begin{figure}[t]
    \centering
    \includegraphics[width=0.45\textwidth]{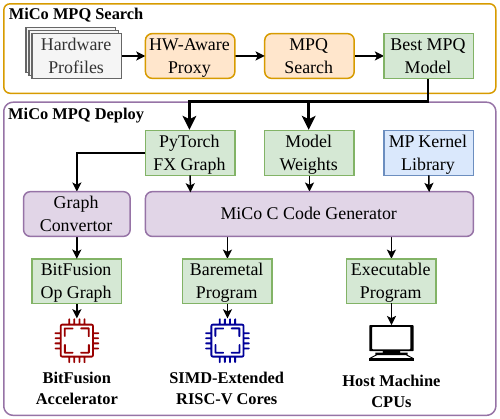}
    \caption{MiCo End-to-End Exploration \& Deployment}
    \label{fig:E2E_ED}
\end{figure}

The framework also offers complete deployment infrastructures, including MPQ model exporting, graph extraction, mixed-precision inference library, and code generation, as shown in Fig.~\ref{fig:E2E_ED}. This creates a seamless transition from PyTorch models to executable bare-metal programs on edge devices.

\subsection{Multi-platform Deployment}

The lower part of Fig.~\ref{fig:E2E_ED} shows the deployment flow of the MiCo framework. The best MPQ model explored by the search flow is first parsed into a PyTorch FX graph, which represents how the model calls neural network functions, operators, and modules during inference. Then, depending on the specific hardware platform to deploy, the graph is converted into different forms that incorporate the MPQ scheme of the model.

Currently, our deployment flow supports 3 hardware platforms: the BitFusion accelerator~\cite{bit_fusion}, the custom SIMD-extended RISC-V CPU, and the host machine that runs the MiCo framework. The core infrastructures of our deployment flow are as follows:

\paragraph{Graph Convertor}
The simulator of BitFusion is based on DnnWeaver~\cite{dnnweaver}. Therefore, to simulate the MPQ model on BitFusion, we have developed a \textbf{graph convertor} that converts the PyTorch FX graph into a DnnWeaver operator graph, and assigns the MPQ scheme to each operation in the graph. This operator graph with bitwidth assignments is deployed and simulated on the BitFusion simulator.

\paragraph{Mixed Precision Computation Library}
For edge devices with CPUs, a bare-metal C code library is essential for the inference of edge AI models. While there are many C libraries like TVM~\cite{TVM} and CMSIS-NN~\cite{cmsis-cnn} that support floating-point and 8-bit integer data, there is limited support for sub-byte data (4-, 2-, and 1-bit). Therefore, we extend the library of baremetal-NN~\cite{baremetal-nn} to support the MPQ kernels (MatMul, Conv2D) with all mixed bitwidth combinations of 8, 4, 2, and 1-bit. Additionally, we have implemented the SIMD-accelerated variants of these MPQ kernels based on VexiiMiCo~\cite{mico_iccad} custom instructions.
To support the dynamic quantization of activations, we have also implemented miscellaneous functions like quantization, de-quantization, packing, and unpacking. With our MP kernel library, the complete inference of MPQ models can be fully supported.


\paragraph{MiCo C Code Generator}
For the deployment on CPUs, the PyTorch FX graph derived from the MPQ model is translated into C code by our C code generator. This generator traces all functions and operations involved in the forward pass of the model, and maps them into the kernels in the aforementioned MP library. Then, it allocates all the buffers for layer inputs and outputs. Finally, it exports the quantized model in binary format, and assigns the addresses to the weight pointers of each layer. The generated C code can be compiled and run on both RISC-V CPUs and the host machine CPUs (e.g., x86).

Additionally, the holistic end-to-end flow is wrapped into Python APIs, which allows us to run the exploration and deployment with only a few lines of Python scripts.


%% file: Content/PTQ_Experiment.tex
\subsection{Mixed Precision PTQ Search Results}

We first evaluate PTQ-based MPQ search over the 4--8-bit space supported by the $w$-based method~\cite{edge-mpq} and BOMP~\cite{van2023bomp}. For all learning-based methods, the search budget comprises 16 initial and 32 search samples. Each experiment is repeated with five random seeds, and the average results are reported in Tab.~\ref{tab:ptq search}.

\input{PTQ_Table}

Our method achieves the best accuracy in most settings. Its advantage is particularly evident for deeper models such as SqueezeNet and TinyLLaMa, whose large search spaces degrade the performance of competing methods. By concentrating candidate generation near the constraint boundary, our method improves sample efficiency while preserving model accuracy.

\begin{figure}[ht]
    \centering
    \includegraphics[width=0.45\textwidth]{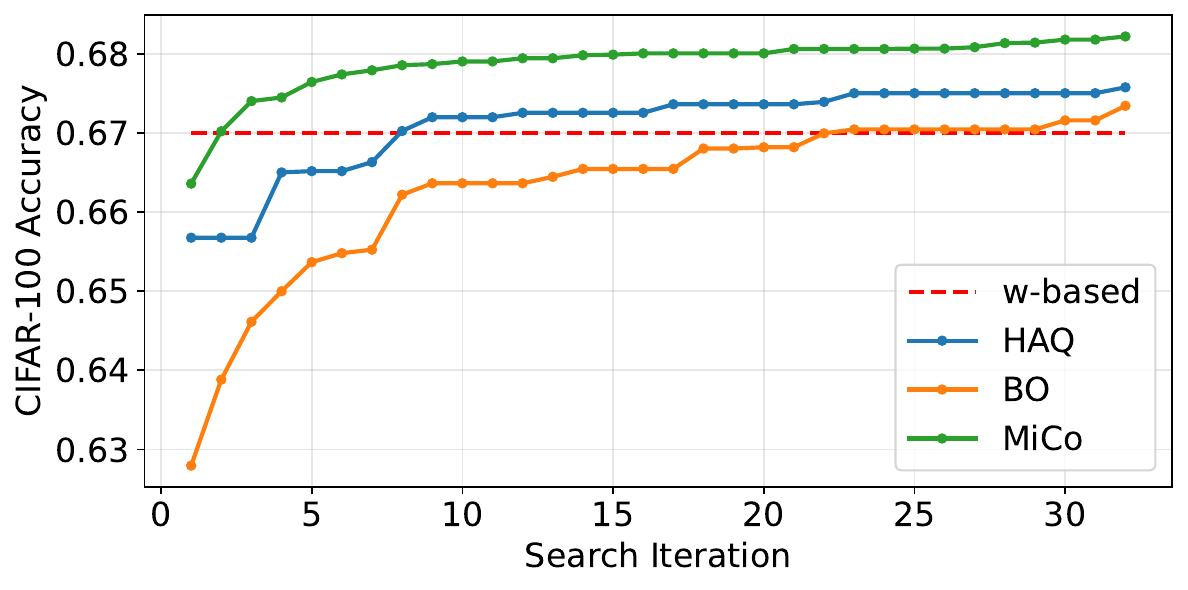}
    \caption{Comparison of PTQ Search Traces on SqueezeNet under $0.6\times$ BOPs}
    \label{fig:ptq_trace}
\end{figure}

Fig.~\ref{fig:ptq_trace} further compares the search traces on SqueezeNet under a $0.6\times$ BOPs constraint. Our method consistently improves the best observed accuracy and surpasses the NLP-based method after only two search iterations, corresponding to 18 total evaluations. In comparison, the NLP-based method requires 26 evaluations to estimate the layer-wise sensitivities.

After 32 search iterations, our method achieves an average accuracy of 68.22\%, compared with 67.58\% for HAQ. HAQ requires 59 total evaluations to reach the same accuracy. Since network evaluation dominates the search time for models such as SqueezeNet, this result demonstrates the higher sample and runtime efficiency of our method.



%% file: PTQ_Table.tex
\begin{table}[ht]
\centering
\caption{PTQ Search (4/5/6/7/8-bit) Accuracy Results}
\label{tab:ptq search}
\begin{tabular}{ l | l | c | c | c }
\hline
\textbf{Model}              & \textbf{Method} & \textbf{0.6BOPs} & \textbf{0.5BOPs} & \textbf{0.4BOPs} \\
\hline
\multirow{4}{*}{CNN4}       & $w$-based       & 74.18 & 73.01  & 74.40 \\
                            & HAQ             & 75.38 & 75.10  & 74.71 \\
                            & BO              & 75.56 & 75.39  & 75.16 \\
                            & \textbf{Ours}   & \textbf{75.61} & \textbf{75.47}  & \textbf{75.21} \\
\hline
\multirow{4}{*}{LeNet5}     & $w$-based       & 99.13 & 99.12 & 99.10 \\
                            & HAQ             & 99.13 & 99.09 & 99.09 \\
                            & BO              & 99.18 & 99.15 & \textbf{99.15} \\
                            & \textbf{Ours}   & \textbf{99.25} & \textbf{99.19} & 99.07 \\
\hline
\multirow{4}{*}{VGG7}       & $w$-based       & 82.84 & 82.75 & 82.73 \\
                            & HAQ             & 83.26 & 83.29 & 83.09 \\
                            & BO              & 83.26 & 83.27 & 83.14 \\
                            & \textbf{Ours}   & \textbf{83.41} & \textbf{83.42} & \textbf{83.24} \\
\hline
\multirow{4}{*}{DS-CNN}     & $w$-based       & 86.08 & 86.42 & 82.96 \\
                            & HAQ             & 88.13 & 86.74 & \textbf{86.56}\\
                            & BO              & 88.55 & \textbf{88.11} & 86.27 \\
                            & \textbf{Ours}   & \textbf{88.63} & 87.43 & 86.39 \\
\hline
\multirow{4}{*}{ResNet-18}  & $w$-based       & 71.34 & 71.20 & 70.51 \\
                            & HAQ             & 71.45 & 71.36 & 71.01\\
                            & BO              & 71.33 & 71.28 & 70.45 \\
                            & \textbf{Ours}   & \textbf{71.49} & \textbf{71.40} & \textbf{71.26} \\
\hline
\multirow{4}{*}{SqueezeNet} & $w$-based       & 67.00 & 66.39 & 63.71 \\
                            & HAQ             & 67.58 & 66.63 & 63.88 \\
                            & BO              & 67.35 & 66.85 & 63.83 \\
                            & \textbf{Ours}   & \textbf{68.22} & \textbf{67.29} & \textbf{65.84}\\
\hline
\multirow{4}{*}{TinyLLaMa-1M}  & $w$-based     & 54.04 & 51.00 & 45.65 \\
                            & HAQ             & 55.60  & 54.63 & \textbf{49.20} \\
                            & BO              & 52.89  & 55.71 & 47.65 \\
                            & \textbf{Ours}   & \textbf{57.20} & \textbf{56.33} & 47.82  \\
\hline
\multirow{4}{*}{ResNet-34}  & $w$-based      & 50.26 & 20.03 & 5.76\\
                            & HAQ            & 65.84 & 58.24 & \textbf{28.48}\\
                            & BO             & 65.56 & 54.95 & 14.76\\
                            & \textbf{Ours}  & \textbf{66.96} & \textbf{61.96} & 20.92 \\
\hline
\multirow{4}{*}{ViT-B-32}  & $w$-based       & 58.61 & 58.61 & 53.60\\
                            & HAQ            & 71.66 & 69.00 & 54.08\\
                            & BO             & 68.57 & 66.74 & \textbf{56.25}\\
                            & \textbf{Ours}  & \textbf{73.63} & \textbf{71.67} & 52.96 \\
\hline
\multirow{4}{*}{TinyLLaMa-7M} & $w$-based     & 56.45 & 54.09 & 44.25 \\
                            & HAQ             & 62.04 & 53.68 & 29.44 \\
                            & BO              & 62.24 & 58.22 & 45.74 \\
                            & \textbf{Ours}   & \textbf{65.05} & \textbf{58.59} & \textbf{47.82}\\
\hline
\end{tabular}
\end{table}

%% file: Content/QAT_Experiment.tex
\subsection{Mixed Precision QAT Search Results}

While many existing MPQ works focus on bit widths above 4 bits (such as 5, 6, 7 bits), the parallelism cannot be effectively enhanced, since most processors and accelerators are designed with arithmetic units that operate with power-of-2 bitwidths (32-bit or 64-bit). For example, if we want to deploy a model with 5-bit weights on most of the hardware in Tab.~\ref{tab:MP_Hardware}, such as XpulpNN~\cite{xpulpnn}, the 5-bit weights need to be extended back to 8-bit before the computation, which hinders the acceleration.

Therefore, to achieve practical speedup in mixed-precision computing, we extend the design space to include bitwidths below 4 bits, which heavily rely on QAT to achieve acceptable accuracy as discussed in Sec.~\ref{sec:ptq_vs_qat}. We conduct QAT search experiments with the bitwidth choices of 1/2/4/8 bits, where each scheme is shortly re-trained during the exploration, and the final best model will be re-trained for an extended period~($\approx$ pre-trained time/10). Due to the longer time consumption and higher computation requirements for QAT, 3 models with larger datasets in Tab.~\ref{tab:networks} are omitted, while the other 7 models are explored.

The search budgets for learning-based methods are set to 32 samples, consisting of 16 initial samples and 16 search samples. We repeat each experiment with 3 different seeds for each algorithm, and report the average results. The QAT search results are listed in Tab.~\ref{tab:qat search}.

\input{QAT_Table}



\color{revise}
As shown in Tab.~\ref{tab:qat search}, QAT effectively recovers the accuracy loss caused by the lower-bitwidth search space. The $w$-based method suffers from directly assigning some layers to 1-bit precision, while BO struggles with the high-dimensional search spaces of larger models. HAQ remains competitive on deeper models, whereas our method achieves the best or comparable accuracy across most models and constraints.
\color{black}

%% file: QAT_Table.tex
\begin{table}[ht]
\centering
\begin{threeparttable}
\caption{QAT Search (1/2/4/8-bit) Accuracy Results}
\label{tab:qat search}
\begin{tabular}{ l | l | c | c | c }
\hline
\textbf{Model}              & \textbf{Method} & \textbf{0.6BOPs} & \textbf{0.5BOPs} & \textbf{0.4BOPs} \\
\hline
\multirow{4}{*}{CNN4}         & $w$-based       & 68.31 & 56.99 & 69.28 \\
                              & HAQ             & 77.86 & 74.01 & 73.68 \\
                              & BO              & 77.27 & 76.16 & 75.75 \\
                              & \textbf{Ours}   & \textbf{78.12} & \textbf{77.34} & \textbf{77.26} \\
\hline
\multirow{4}{*}{LeNet5}       & $w$-based       & 99.22 & 98.28 & 99.28 \\
                              & HAQ             & 99.26 & 99.21 & 99.20 \\
                              & BO              & 99.03 & 99.03 & 99.07 \\
                              & \textbf{Ours}   & \textbf{99.32} & \textbf{99.25} & \textbf{99.36} \\
\hline
\multirow{4}{*}{VGG7}         & $w$-based       & 84.90 & 70.49 & 80.35 \\
                              & HAQ             & 87.31 & \textbf{87.67} & 87.33 \\
                              & BO              & 84.39 & 84.10 & 84.01 \\
                              & \textbf{Ours}   & \textbf{87.74} & 87.60 & \textbf{87.74} \\
\hline
\multirow{4}{*}{ResNet-18}   & $w$-based        & 69.16 & 62.03 & 61.72 \\
                             & HAQ  & \textbf{70.65} & \textbf{69.96} & 68.40 \\
                            & BO              & 59.16 & 59.13 & 59.18 \\
                             & \textbf{Ours}   & 69.74 & 69.62 & \textbf{68.46} \\
\hline
\multirow{4}{*}{SqueezeNet}   & $w$-based       & 62.95 & 61.57 & 61.08 \\
                              & HAQ             & 64.52 & 64.45 & 60.21 \\
                              & BO              & 42.85 & 43.65 & 43.82 \\
                              & \textbf{Ours}   & \textbf{65.24} & \textbf{64.66} & \textbf{62.46} \\
\hline
\multirow{4}{*}{TinyLLaMa-1M} & $w$-based       & 48.03 & 47.93 & 42.29 \\
                              & HAQ             & 50.55 & 51.43 & \textbf{50.70} \\
                              & BO              & 33.54 & 35.29 & 35.34 \\
                              & \textbf{Ours}   & \textbf{53.45} & \textbf{51.79} & 49.75 \\
\hline
\multirow{4}{*}{TinyLLaMa-7M} & $w$-based       & 51.51 & 49.77 & 39.80 \\
                              & HAQ             & 58.23 & \textbf{57.05} & 50.27 \\
                              & BO              & 43.30 & 43.27 & 43.39 \\
                              & \textbf{Ours}   & \textbf{59.12} & 54.76 & \textbf{52.61} \\
\hline
\end{tabular}
\end{threeparttable}
\end{table}

%% file: Content/NewExperiment.tex
\color{revise}
\subsection{Comparison with Additional Baselines}
\label{sec:uniform_baseline}
To further validate our approach, we evaluate uniform and
empirical policies, and the ILP-based HAWQ-V3~\cite{hawqV3}, on
ResNet-18 (PTQ, 0.6 BOPs) and VGG7 (QAT, 0.4 BOPs), as listed in
Tab.~\ref{tab:uniform_baseline}. Except for the W8A8 reference, which exceeds the constraint, all schemes satisfy their respective BOPs constraints (21.70G for ResNet-18, 2.84G for VGG7).
MiCo achieves the highest accuracy on both models. While the heuristic of keeping the first and last layers at 8-bit yields competitive results, our layer-wise mixed-precision search achieves further improvement without exceeding the constraints.
\color{black}

\begin{table}[htbp]
\centering
\color{revise}
\footnotesize
\begin{threeparttable}
\caption{Comparison with Additional Baselines}
\label{tab:uniform_baseline}
\begin{tabular}{l|cc|cc}
\hline
\multirow{2}{*}{\textbf{Method}} & \multicolumn{2}{c|}{\textbf{ResNet-18} (PTQ, 0.6)} & \multicolumn{2}{c}{\textbf{VGG7} (QAT, 0.4)} \\
\cline{2-5}
 & Acc.~(\%) & BOPs & Acc.~(\%) & BOPs \\
\hline
W8A8                 & 71.48 & 36.16G & 87.72 & 7.10G \\
\hline
W6A6/W4A4\tnote{a}   & 70.95 & 20.34G & 86.14 & 1.77G \\
First/Last 8-bit\tnote{b}    & 71.40 & 20.39G & 87.34 & 1.86G \\
HAWQ-V3~\cite{hawqV3} & 70.93 & 19.53G & 86.21 & 2.57G \\
$w$-based~\cite{edge-mpq} & 71.34 & 21.33G & 80.35 & 2.75G \\
HAQ~\cite{haq}        & 71.45 & 21.62G & 87.33 & 2.33G \\
BO~\cite{van2023bomp} & 71.33 & 19.58G & 84.01 & 2.80G \\
\textbf{MiCo}         & \textbf{71.49} & 20.46G & \textbf{87.74} & 2.73G \\
\hline
\end{tabular}
\begin{tablenotes}
\item [a] W6A6 for ResNet-18, W4A4 for VGG7.
\item [b] Same as (a) except the first and last layers are kept in 8-bit.
\end{tablenotes}
\end{threeparttable}
\end{table}

%% file: ProxyTable.tex
\begin{table}[htbp]
\centering
\caption{Hardware-Aware Proxy for Different Networks on BitFusion}
\label{tab:proxy_bf}
\begin{tabular}{l|cc|cc}
\hline
\multirow{2}{*}{Features} & \multicolumn{2}{c|}{VGG7 on BitFusion} & \multicolumn{2}{c}{ResNet-18 on BitFusion} \\
\cline{2-5}
    & MAPE$\downarrow$ & $R^2\uparrow$   & MAPE$\downarrow$ & $R^2\uparrow$      \\
\hline
BOPs     & 9.35         & 0.50                 & 5.68           & 0.70                    \\
CBOPs    & 4.96         & 0.88                 & 3.46           & 0.78                    \\
\textcolor{revise}{Full Feat.} & 2.76         & 0.94                 & 1.66           & 0.80                   \\
\hline
\end{tabular}
\end{table}

\begin{table}[htbp]
\centering
\caption{Hardware-Aware Proxy for CNN4 on Different VexiiMiCo Variants}
\label{tab:proxy_mico}
\begin{tabular}{l|cc|cc|cc}
\hline
\multirow{2}{*}{Features} & \multicolumn{2}{c|}{\textit{Tiny}} & \multicolumn{2}{c|}{\textit{Small}} & \multicolumn{2}{c}{\textit{High}} \\
\cline{2-7}
 & MAPE$\downarrow$ & $R^2\uparrow$ & MAPE$\downarrow$ & $R^2\uparrow$ & MAPE$\downarrow$ & $R^2\uparrow$ \\
\hline
BOPs     & 20.47 & -0.92 & 19.66        & -1.64        & 23.31          & -0.97           \\
CBOPs    & 6.95  & 0.75  & 6.32         & 0.77         & 7.38           & 0.75            \\
\textcolor{revise}{Full Feat.} & 4.78  & 0.85  & 3.24         & 0.89         & 3.59           & 0.96            \\      
\hline
\end{tabular}
\end{table}

%% file: TransferTable.tex
\newcommand{\greenL}{\cellcolor[HTML]{44FF44}}
\newcommand{\greenM}{\cellcolor[HTML]{CCFFCC}}
\newcommand{\lightyellow}{\cellcolor[HTML]{FFFFCC}}
\newcommand{\lightorange}{\cellcolor[HTML]{FFE5CC}}

\begin{table}[htbp]
\centering
\caption{Transfer Learning Between VexiiMiCo Variants}
\label{tab:transfer_learning}
\setlength{\tabcolsep}{4pt}
\footnotesize
\begin{tabular}{l|cc|cc|cc}
\hline
 \multirow{2}{*}{Method} & \multicolumn{2}{c|}{\textit{Tiny}} 
 & \multicolumn{2}{c|}{\textit{Small}} 
 & \multicolumn{2}{c}{\textit{High}} \\
 \cline{2-7}
       & MAPE$\downarrow$ & $R^2\uparrow$
       & MAPE$\downarrow$ & $R^2\uparrow$
       & MAPE$\downarrow$ & $R^2\uparrow$ \\
\hline
Direct Fit (5\%)  & \lightyellow 14.49 & \lightyellow -0.96
            & \lightyellow 14.68 & \lightyellow -1.84 
            & \lightyellow 19.10 & \lightyellow -1.38 \\
\hline
TL from Tiny      & --   & --     
                  & \greenM 4.62 & \greenM 0.82   
                  & \greenM 7.09 & \greenM 0.89 \\
TL from Small     & \greenM 4.97 & \greenM 0.65  
                  & --   & --     
                  & \greenL 5.70 & \greenL 0.93 \\
TL from High      & \greenL 4.63 & \greenL 0.87   
                  & \greenL 3.87 & \greenL 0.84   
                  & --   & --   \\
\hline
\end{tabular}
\end{table}

%% file: vgg_bitwidths.tex
    \centering
    \scalebox{0.85}{%
    \begin{tikzpicture}
    \begin{axis}[
        ybar,
        bar shift=0pt, 
        bar width=8pt, 
        width=\linewidth, 
        height=5cm, 
        xlabel={\textbf{Network Layer}},
        ylabel={\textbf{Precision (Bits)}},
        symbolic x coords={1, 2, 3, 4, 5, 6, 7},
        xtick=data,
        ymin=-10, ymax=10, 
        ytick={-8, -6, -4, -2, 0, 2, 4, 6, 8},
        yticklabels={8, 6, 4, 2, 0, 2, 4, 6, 8}, 
        ymajorgrids=true,                   
        major grid style={dashed, gray!40}, 
        axis on top=false,                  
        legend style={at={(0.5, 1.2)}, anchor=north, legend columns=-1, draw=gray!50, font=\small},
        extra y ticks={0},
        extra y tick style={grid=major, major grid style={thick, black}},
        enlarge x limits=0.15,
    ]

    \addplot[
        fill=blue!50, 
        draw=blue!80!black, 
        nodes near coords, 
        nodes near coords align={vertical},
        font=\small
    ] coordinates {
        (1, 8) 
        (2, 7) 
        (3, 6) 
        (4, 7) 
        (5, 4) 
        (6, 4) 
        (7, 8)
    };

    \addplot[
        fill=orange!70, 
        draw=orange!80!black, 
        nodes near coords, 
        nodes near coords align={below},
        point meta=explicit,
        font=\small
    ] coordinates {
        (1, -8) [8]
        (2, -4) [4]
        (3, -8) [8]
        (4, -4) [4]
        (5, -6) [6]
        (6, -8) [8]
        (7, -8) [8]
    };

    \legend{Weight Precision, Activation Precision}

    \end{axis}
    \end{tikzpicture}
    }
    \vspace{-0.1in}
    \caption{VGG7}
    \label{fig:vgg-bitwidth}

%% file: resnet18_bitwidths.tex
    \centering
    \scalebox{0.85}{%
    \begin{tikzpicture}
    \begin{axis}[
        ybar,
        bar shift=0pt,
        bar width=8pt, 
        width=\linewidth, 
        height=5cm,
        xlabel={\textbf{Network Layer}},
        ylabel={\textbf{Precision (Bits)}},
        symbolic x coords={1, 2, 3, 4, 5, 6, 7, 8, 9, 10, 11, 12, 13, 14, 15, 16, 17, 18, 19, 20, 21},
        xtick=data,
        ymin=-10, ymax=10, 
        ytick={-8, -6, -4, -2, 0, 2, 4, 6, 8},
        yticklabels={8, 6, 4, 2, 0, 2, 4, 6, 8},
        ymajorgrids=true,
        major grid style={dashed, gray!40},
        axis on top=false,
        legend style={at={(0.5, 1.2)}, anchor=north, legend columns=-1, draw=gray!50, font=\small},
        extra y ticks={0},
        extra y tick style={grid=major, major grid style={solid, thick, black}}, 
        enlarge x limits=0.05, 
    ]
    \addplot[fill=blue!50, draw=blue!80!black] coordinates {
      (1,8) (2,8) (3,5) (4,7) (5,4) (6,7) (7,7) (8,6) (9,6) (10,4) (11,6) 
      (12,4) (13,6) (14,5) (15,6) (16,5) (17,4) (18,8) (19,6) (20,5) (21,8)
    };
    \addplot[fill=orange!70, draw=orange!80!black] coordinates {
      (1,-8) (2,-5) (3,-4) (4,-4) (5,-4) (6,-4) (7,-4) (8,-8) (9,-4) (10,-4) (11,-8) 
      (12,-4) (13,-7) (14,-6) (15,-4) (16,-7) (17,-4) (18,-4) (19,-4) (20,-8) (21,-8)
    };
    \legend{Weight Precision, Activation Precision}
    \end{axis}
    \end{tikzpicture}
    }
    \vspace{-0.1in}
    \caption{ResNet-18}
    \label{fig:resnet18-bitwidth}

%% file: Content/Result_On_CPUs.tex
\subsection{End-to-End Exploration on SIMD-Extended RISC-V CPUs}
\label{sec:e2e_on_rv}
In this experiment, we utilize SIMD-extended RISC-V CPUs~(VexiiMiCo) as the hardware targets to demonstrate our exploration-deployment framework.

We perform MPQ exploration on the CNN4 and LeNet5 in Tab.~\ref{tab:networks}, deploy MPQ models on the target CPUs, and collect cycle-accurate execution latencies via RTL simulations. As explained in previous sections, we adopt the QAT flow to recover the accuracy for the bitwidth space that includes 1-/2-bit choices. 

\begin{table}[ht]
\centering
\caption{Exploration \& Deployment on VexiiMiCo CPUs}
\label{tab:mico_cpu}
\begin{tabular}{l|l|lll}
\hline
\multicolumn{5}{c}{\textbf{CNN4 (CIFAR-10)}} \\
\hline
\textbf{Type} & \textbf{Precision} &\textbf{Acc.} (\%) & \textbf{Constraint} & \textbf{Cycles} (Ratio) \\
\hline
\multirow{3}{*}{\textit{Tiny}} & 8-bit (QAT)  & 78.93 & -        &  68.0M (1.0$\times$)  \\
& Mixed (QAT) & 78.95  & 0.8$\times$BOPs   &  63.0M (0.93$\times$) \\
& Mixed (QAT) & 78.24  & 0.8$\times$\textit{Proxy}  &  \textbf{54.7M (0.80$\times$)}\\
\hline
\multirow{3}{*}{\textit{Small}} & 8-bit (QAT)  & 78.93 & -        & 64.4M (1.0$\times$)  \\
& Mixed (QAT) & 78.95 & 0.8$\times$BOPs   & 60.0M (0.93$\times$) \\
& Mixed (QAT) & 78.75 & 0.8$\times$\textit{Proxy}  & \textbf{55.1M (0.85$\times$)} \\
\hline
\multirow{3}{*}{\textit{High}} & 8-bit (QAT)  & 78.93 & -        & 41.8M (1.0$\times$)  \\
& Mixed (QAT) & 78.95 & 0.8$\times$BOPs  & 40.2M (0.97$\times$) \\
& Mixed (QAT) & 78.76 & 0.8$\times$\textit{Proxy} & \textbf{32.7M (0.78$\times$)} \\
\hline
\multicolumn{5}{c}{\textbf{LeNet5 (MNIST)}} \\
\hline
\textbf{Type} & \textbf{Precision} &\textbf{Acc.} (\%) & \textbf{Constraint} & \textbf{Cycles} (Ratio) \\
\hline
\multirow{3}{*}{\textit{Tiny}} & 8-bit (QAT)  &  99.35 & -        &  4.45M (1.0$\times$)  \\
& Mixed (QAT) & 99.35 & 0.8$\times$BOPs  &  4.19M (0.94$\times$) \\
& Mixed (QAT) & 96.90 & 0.8$\times$\textit{Proxy} &  \textbf{3.51M (0.79$\times$)}\\
\hline
\multirow{3}{*}{\textit{Small}} & 8-bit (QAT)&  99.35 & -        & 4.03M (1.0$\times$)  \\
& Mixed (QAT) & 99.35 & 0.8$\times$BOPs                      & 3.94M (0.98$\times$) \\
& Mixed (QAT) & 96.90 & 0.8$\times$\textit{Proxy}            & \textbf{3.45M (0.85$\times$)} \\
\hline
\multirow{3}{*}{\textit{High}} & 8-bit (QAT) &  99.35   & - & 2.67M (1.0$\times$)  \\
& Mixed (QAT) & 99.35 & 0.8$\times$BOPs           & 2.60M (0.98$\times$) \\
& Mixed (QAT) & 96.46 & 0.8$\times$\textit{Proxy} & \textbf{2.22M (0.83$\times$)} \\
\hline
\end{tabular}
\end{table}

As shown in Tab.~\ref{tab:mico_cpu}, although optimizing with BOPs constraints can lead to speed up as well, the improvement is much lower than expected. This is highly related to the greater non-linearity of CPU-based inference compared to accelerators, due to factors like caching and operations such as im2col or dynamic quantization. With hardware-aware proxy models for each CPU configuration, the constraints successfully guide the exploration towards higher speedup, leading to lower end-to-end latency for the MPQ model deployment.